\documentclass[conference]{IEEEtran}
\IEEEoverridecommandlockouts
\usepackage{xcolor} 
\usepackage{array}
\usepackage{amsfonts}
\usepackage{graphicx}

\usepackage{epstopdf}
\usepackage{amssymb}
\usepackage{tikz,pgfplots}
\usetikzlibrary{snakes,arrows,shapes,trees}
\usepackage{amsopn}
\usepackage{listings}
\usepackage{adjustbox}
\usepackage{longtable}
\usepackage{multirow}
\usepackage{hyperref}
\usepackage{amssymb}
\usepackage{pgfplots}
\usepackage{pgfplotstable}
\usetikzlibrary{arrows,shapes,plotmarks}
\pgfplotsset{compat=1.8}
\usepackage{soul}
\usepackage{rotating}
\usepackage{url}
\usepackage{algorithm}
\usepackage{algpseudocode}
\usepackage{enumitem}
\usepackage{caption}
\usepackage{subcaption}
\usepackage{csvsimple}
\usepackage{makecell}
\usepackage[]{footmisc}

\hypersetup{
	colorlinks,
	linkcolor={blue!50!black},
	citecolor={blue!50!black},
	urlcolor={blue!80!black},
	anchorcolor = {blue!80!black},
	filecolor = {blue!80!black},
	menucolor = {blue!80!black},
	runcolor = {blue!80!black}
}

\newdimen\HilbertLastX
\newdimen\HilbertLastY
\newcounter{HilbertOrder}

\def\DrawToNext#1#2{%
  \advance \HilbertLastX by #1
  \advance \HilbertLastY by #2
  \pgfpathlineto{\pgfqpoint{\HilbertLastX}{\HilbertLastY}}
}

\def\Hilbert[#1,#2,#3,#4,#5,#6,#7,#8] {
  \ifnum\value{HilbertOrder} > 0%
  \addtocounter{HilbertOrder}{-1}
  \Hilbert[#5,#6,#7,#8,#1,#2,#3,#4]
  \DrawToNext {#1} {#2}
  \Hilbert[#1,#2,#3,#4,#5,#6,#7,#8]
  \DrawToNext {#5} {#6}
  \Hilbert[#1,#2,#3,#4,#5,#6,#7,#8]
  \DrawToNext {#3} {#4}
  \Hilbert[#7,#8,#5,#6,#3,#4,#1,#2]
  \addtocounter{HilbertOrder}{1}
  \fi
}

\def\hilbert((#1,#2),#3){%
  \advance \HilbertLastX by #1
  \advance \HilbertLastY by #2
  \pgfpathmoveto{\pgfqpoint{\HilbertLastX}{\HilbertLastY}}
  \setcounter{HilbertOrder}{#3}
  \Hilbert[1mm,0mm,-1mm,0mm,0mm,1mm,0mm,-1mm]
  \pgfusepath{stroke}%
}

\ifpdf
  \DeclareGraphicsExtensions{.eps,.pdf,.png,.jpg}
\else
  \DeclareGraphicsExtensions{.eps}
\fi

\definecolor{cpu3}{HTML}{F44336}
\definecolor{cpu4}{HTML}{2196F3}
\definecolor{cpu1}{HTML}{4CAF50}
\definecolor{cpu2}{HTML}{FFC107}
\definecolor{gpu3}{HTML}{EF9A9A}
\definecolor{gpu4}{HTML}{90CAF9}
\definecolor{gpu1}{HTML}{A5D6A7}
\definecolor{gpu2}{HTML}{FFE082}

\definecolor{cpu5}{HTML}{9932CC}

\definecolor{sq_b1}{RGB}{37,52,148}
\definecolor{sq_b2}{RGB}{44,127,184}
\definecolor{sq_b3}{RGB}{65,182,196}
\definecolor{sq_b4}{RGB}{127,205,187}
\definecolor{sq_b5}{RGB}{199,233,180}
\definecolor{sq_b6}{RGB}{255,255,204}

\definecolor{sq_r1}{RGB}{189,0,38}
\definecolor{sq_r2}{RGB}{240,59,32}
\definecolor{sq_r3}{RGB}{253,141,60}
\definecolor{sq_r4}{RGB}{254,178,76}
\definecolor{sq_r5}{RGB}{254,217,118}
\definecolor{sq_r6}{RGB}{255,255,178}

\definecolor{sq_g1}{RGB}{0,104,55}
\definecolor{sq_g2}{RGB}{49,163,84}
\definecolor{sq_g3}{RGB}{120,198,121}
\definecolor{sq_g4}{RGB}{173,221,142}
\definecolor{sq_g5}{RGB}{217,240,163}
\definecolor{sq_g6}{RGB}{255,255,204}

\definecolor{div_c1}{RGB}{230,171,2}
\definecolor{div_c2}{RGB}{102,166,30}
\definecolor{div_c3}{RGB}{231,41,138}
\definecolor{div_c4}{RGB}{117,112,179}
\definecolor{div_c5}{RGB}{217,95,2}
\definecolor{div_c6}{RGB}{27,158,119}
\definecolor{div_c7}{RGB}{215,48,39}

\definecolor{div_d1}{RGB}{215,25,28}
\definecolor{div_d2}{RGB}{253,174,97}
\definecolor{div_d3}{RGB}{255,255,191}
\definecolor{div_d4}{RGB}{171,217,233}
\definecolor{div_d5}{RGB}{44,123,182}

\definecolor{lineclr}{RGB}{0,0,0}
\definecolor{utorange}{RGB}{0,0,255}
\definecolor{utsecblue}{RGB}{255,255,0}
\definecolor{utsecgreen}{RGB}{255,0,0}
\definecolor{red!15}{RGB}{0,255,255}
\definecolor{fillclr5}{RGB}{0,255,0}
\definecolor{fillclr6}{RGB}{255,0,255}
\definecolor{fillclr7}{RGB}{255,255,255}
\definecolor{fillclr8}{RGB}{0,0,0}

\def\drawcubeI(#1,#2,#3,#4,#5){ 
\coordinate (O) at (#1,#2,#3);
\coordinate (A) at (#1,#2+#4,#3);
\coordinate (B) at (#1,#2+#4,#3+#4);
\coordinate (C) at (#1,#2,#3+#4);
\coordinate (D) at (#1+#4,#2,#3);
\coordinate (E) at (#1+#4,#2+#4,#3);
\coordinate (F) at (#1+#4,#2+#4,#3+#4);
\coordinate (G) at (#1+#4,#2,#3+#4);
\draw[#5] (O) -- (C) -- (G) -- (D) -- cycle;
\draw[#5] (O) -- (A) -- (E) -- (D) -- cycle;
\draw[#5] (O) -- (A) -- (B) -- (C) -- cycle;
\draw[#5] (D) -- (E) -- (F) -- (G) -- cycle;
\draw[#5] (C) -- (B) -- (F) -- (G) -- cycle;
\draw[#5] (A) -- (B) -- (F) -- (E) -- cycle;
}

\def\drawcubeII(#1,#2,#3,#4,#5,#6,#7){ 
\coordinate (O) at (#1,#2,#3);
\coordinate (A) at (#1,#2+#4,#3);
\coordinate (B) at (#1,#2+#4,#3+#4);
\coordinate (C) at (#1,#2,#3+#4);
\coordinate (D) at (#1+#4,#2,#3);
\coordinate (E) at (#1+#4,#2+#4,#3);
\coordinate (F) at (#1+#4,#2+#4,#3+#4);
\coordinate (G) at (#1+#4,#2,#3+#4);
\draw[#5,fill=#6,opacity=#7] (O) -- (C) -- (G) -- (D) -- cycle;
\draw[#5,fill=#6,opacity=#7] (O) -- (A) -- (E) -- (D) -- cycle;
\draw[#5,fill=#6,opacity=#7] (O) -- (A) -- (B) -- (C) -- cycle;
\draw[#5,fill=#6,opacity=#7] (D) -- (E) -- (F) -- (G) -- cycle;
\draw[#5,fill=#6,opacity=#7] (C) -- (B) -- (F) -- (G) -- cycle;
\draw[#5,fill=#6,opacity=#7] (A) -- (B) -- (F) -- (E) -- cycle;
}

\def\drawNodes(#1,#2,#3,#4,#5,#6,#7){ 
\foreach \x in {#1,#7,...,#2}{
	\foreach \y in {#3,#7,...,#4}{
		\foreach \z in {#5,#7,...,#6}{
				\draw[fill=red!60] (\x,\y,\z) circle (0.15);
				}
			}
	}				
		
}

\makeatletter
\newcommand\resetstackedplots{
\makeatletter
\pgfplots@stacked@isfirstplottrue
\makeatother
\addplot [forget plot,draw=none] coordinates{(48,0) (96,0) (192,0) (384,0) (768,0) (1536,0) (3072,0) (6144,0)};
}
\makeatother

\makeatletter
\newcommand\resetstackedplotsOne{
\makeatletter
\pgfplots@stacked@isfirstplottrue
\makeatother
\addplot [forget plot,draw=none] coordinates{(384,0) (768,0) (1536,0) (3072,0) (6144,0)};
}
\makeatother

\makeatletter
\newcommand\resetstackedplotsTwo{
\makeatletter
\pgfplots@stacked@isfirstplottrue
\makeatother
\addplot [forget plot,draw=none] coordinates{(16,0) (32,0) (64,0) (128,0) (256,0) (512,0) (1024,0) (2048,0) (4096,0) (8192,0) (16384,0) (32768,0)};
}
\makeatother

\makeatletter
\newcommand\resetstackedplotsThree{
\makeatletter
\pgfplots@stacked@isfirstplottrue
\makeatother
\addplot [forget plot,draw=none] coordinates{(2,0) (4,0) (8,0) (16,0) (32,0) (64,0)};
}
\makeatother

\makeatletter
\newcommand\resetstackedplotsFour{
\makeatletter
\pgfplots@stacked@isfirstplottrue
\makeatother
\addplot [forget plot,draw=none] coordinates{(4,0) (8,0) (16,0) (32,0) (64,0)};
}
\makeatother

\makeatletter
\newcommand\resetstackedplotsFive{
\makeatletter
\pgfplots@stacked@isfirstplottrue
\makeatother
\addplot [forget plot,draw=none] coordinates{(1,0) (2,0) (4,0) (8,0) (16,0) (32,0) (64,0) (128,0)};
}
\makeatother

\makeatletter
\newcommand\resetstackedplotsSix{
\makeatletter
\pgfplots@stacked@isfirstplottrue
\makeatother
\addplot [forget plot,draw=none] coordinates{(2,0) (4,0) (8,0) (16,0) (32,0) (64,0) (128,0)};
}
\makeatother

\newcommand{\ipoint}[1]{{\color{black}{#1}}}

\pdfoutput=1

\usepackage{amsmath,amsfonts,bm, amsthm,algorithm, algpseudocode}

\def\Figref#1{Fig.~\ref{#1}}

\def\eqref#1{eq.~\ref{#1}}

\def\ceil#1{\lceil #1 \rceil}
\def\floor#1{\lfloor #1 \rfloor}
\def\1{\bm{1}}

\def\rvb{{\mathbf{b}}}

\def\rvu{{\mathbf{i}}}

\def\rvu{{\mathbf{u}}}

\def\rvx{{\mathbf{x}}}

\def\rmA{{\mathbf{A}}}

\def\rmS{{\mathbf{S}}}

\DeclareMathAlphabet{\mathsfit}{\encodingdefault}{\sfdefault}{m}{sl}
\SetMathAlphabet{\mathsfit}{bold}{\encodingdefault}{\sfdefault}{bx}{n}

\def\gA{{\mathcal{A}}}
\def\gB{{\mathcal{B}}}

\def\sD{{\mathbb{D}}}

\def\sR{{\mathbb{R}}}

\newcommand{\etal}{\emph{et. al.}}

\newcommand{\E}{\mathbb{E}}

\theoremstyle{plain}

\theoremstyle{remark}

\theoremstyle{definition}

\theoremstyle{plain}

\theoremstyle{plain}

\theoremstyle{definition}

\providecommand{\corollaryname}{Corollary}
\providecommand{\lemmaname}{Lemma}
\providecommand{\problemname}{Problem}
\providecommand{\remarkname}{Remark}
\providecommand{\theoremname}{Theorem}

\def\BibTeX{{\rm B\kern-.05em{\sc i\kern-.025em b}\kern-.08em
    T\kern-.1667em\lower.7ex\hbox{E}\kern-.125emX}}
\begin{document}
\title{Deep Generative Models that Solve PDEs: Distributed Computing for Training Large Data-Free Models
}

\author{\IEEEauthorblockN{\hspace{1.5cm} Sergio Botelho\textsuperscript{\textsection}
\hspace{1.5cm}}
\IEEEauthorblockA{
\textit{RocketML Inc.}\\
sergio@rocketml.net}
\and
\IEEEauthorblockN{\hspace{1.2cm}Ameya Joshi\textsuperscript{\textsection}\hspace{1.5cm}}
\IEEEauthorblockA{
\textit{New York University (NYU)}\\
ameya.joshi@nyu.edu}
\and
\IEEEauthorblockN{\hspace{2cm}Biswajit Khara\textsuperscript{\textsection}\hspace{2cm}}
\IEEEauthorblockA{
\textit{Iowa State University}\\
bkhara@iastate.edu}
\and
\IEEEauthorblockN{Soumik Sarkar}
\IEEEauthorblockA{
\textit{Iowa State University}\\
soumiks@iastate.edu}
\and
\IEEEauthorblockN{Chinmay Hegde}
\IEEEauthorblockA{
\textit{New York University (NYU)}\\
chinmay.h@nyu.edu}
\and
\IEEEauthorblockN{Santi Adavani}
\IEEEauthorblockA{
\textit{RocketML Inc}\\
santi@rocketml.net}
\and
\IEEEauthorblockN{Baskar Ganapathysubramanian}
\IEEEauthorblockA{
\textit{Iowa State University}\\
baskarg@iastate.edu}
}

\maketitle
\begin{abstract}
Recent progress in scientific machine learning (SciML) has opened up the possibility of training novel neural network architectures that solve complex partial differential equations (PDEs). Several (nearly data free) approaches have been recently reported that successfully solve PDEs, with examples including deep feed forward networks, generative networks, and deep encoder-decoder networks. However, practical adoption of these approaches is limited by the difficulty in training these models, especially to make predictions at large output resolutions ($\geq 1024 \times 1024$). 

Here we report on a software framework for data parallel distributed deep learning that resolves the twin challenges of training these large SciML models – training in reasonable time as well as distributing the storage requirements. Our framework provides several out of the box functionality including (a) loss integrity independent of number of processes, (b) synchronized batch normalization, and (c) distributed higher-order optimization methods.

We show excellent scalability of this framework on both cloud as well as HPC clusters, and report on the interplay between bandwidth, network topology and bare metal vs cloud. We deploy this approach to train generative  models  of  sizes  hitherto  not  possible, showing that neural PDE solvers can be viably trained for practical applications. We also demonstrate that distributed higher-order optimization methods are 2-3$\times$ faster than stochastic gradient-based methods and provide minimal convergence drift with higher batch-size.

\end{abstract}

{\let\thefootnote\relax\footnote{{\textsuperscript{\textsection}Equal contribution}}}


\section{Introduction}

Numerical simulation is a critical tool in analysis, optimization, design, and control of complex engineered systems. The status quo has predominantly been describing and modeling of such systems through partial differential equations (PDEs) and their numerical approximations. For increasingly complex engineered applications (aircraft, rockets, autonomous systems, etc.) the availability of fast predictive models becomes critical, especially if the intent is to use these models for design and/or control (so called model-predictive control, MPC).

Modern deep learning approaches have transformed a host of application areas that involve assimilating large data streams to make useful predictions. There has been increasing interest in leveraging these advances for analysis, optimization, design and control of complex engineered systems~(\cite{paganini2018calogan, king2018deep, chang2016study, pun2018physically, de2017learning, sanchez2018inverse}). However, off-the-shelf utilization of deep learning strategies have had limited applicability, primarily due to the following drawbacks: 
\begin{itemize}
    \item \textit{Reliance on abundance of data}: Current ML approaches tend to entirely let data dictate the narrative. As a result, the data requirements for training such systems is very large, which may be a major bottleneck for complex simulations;
    \item \textit{Lack of generalizability}: They are of narrow scope, i.e., they typically only succeed on the task that they are trained on. Additionally, contextual constraints and domain knowledge known from physical system are left unused.  
 \end{itemize}
 These key issues have motivated the development of \textit{\textbf{Scientific Machine Learning}} (SciML) strategies that seek to bridge modern deep learning concepts with numerical solutions of PDE's. Recent very exciting advances~(\cite{yang2018physics, raissi2017physics1, raissi2018hidden, HsiehPDE2019, joshigenerative}) have shown the efficacy of deep networks in solving partial differential equations (PDEs). 
 Specifically, methods as described in~\cite{HsiehPDE2019, joshigenerative} rely on convolutional neural networks as a \emph{natural} representation of the domain for a PDE. The reliance on data is reduced by explicitly incorporating notions of symmetry, invariance or constraints into the network (either in the loss function, or in the network definition). This also enables better generalizability (due to the satisfaction of the constraints). By training a deep neural network to act as (an arbitrarily accurate) surrogate for a PDE (either a specific instance, or a class of PDEs), significant gains in computational speed have been shown to be possible. This is because the inference stage of neural networks is (near) real time, compared to the cost of training. \textcolor{black}{Thus, given a trained network that acts as a PDE solver for arbitrary boundary and initial conditions, the time-to-solve from a practitioner perspective is simply the time for inference. This is consistent with ML standard practices, where the (non-trivial) cost of training is amortized over the large number of inferences required in, say, model predictive control of complex systems.} 
 
\begin{figure}[htp]
\centering
\captionsetup{justification=centering}
          \centering
      \includegraphics[trim=0 0 0 0,clip,width=1.0\linewidth]{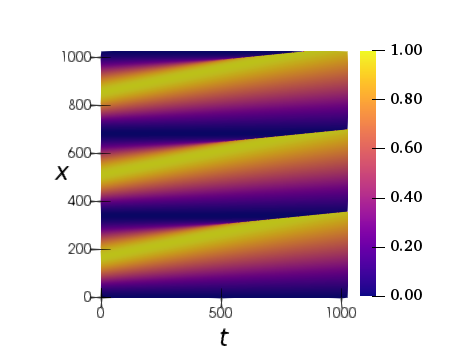}
      \caption{A solution to the inviscid Burgers' equation showing shock formation, solved through finite element method formulated in full space-time domain. \color{black}Physically $(x,t)$ $\in$ $[0,1]\times[0,0.2]$ but all contour plots are rendered in a scaled square grid}
      \label{fig:fe-m1024-c03}
\end{figure}

 While this field is evolving very rapidly, a preliminary taxonomy of `\textit{neural-PDE}' approaches (through the lens of computational science) is as follows: (a) \textit{\textbf{PDE instance vs PDE family solvers}}: Some approaches focus on improving the numerical linear algebra (\cite{HsiehPDE2019, pang2018fpinns, Long2017PDENetLP, Greenfeld2019LearningTO, katrutsa2017deep}), and are limited to a single instance of a PDE, while other strategies (\cite{ yang2018physics, joshigenerative, zhu2019physics, sirignano2018dgm}) focus on solving a general class of PDEs. Instance solvers have the advantage of excellent performance, but need to be retrained for each problem realization; (b) \textit{\textbf{point-wise predictions vs full field predictions}}: Some approaches focus on making point wise predictions in the domain (\cite{ yang2018physics, raissi2017physics1, raissi2018hidden, pang2018fpinns, karumuri2020simulator, han2018solving, michoski2019solving}), while others~(\cite{paganini2018calogan,  joshigenerative, zhu2019physics}) make full field predictions. Point-wise predictions have the advantage of easier trainability (since the output is usually a single scalar), but full-field predictions naturally account for boundary conditions.  \textit{A common bottleneck to these 'neural-PDE' approaches is that nearly all of them scale poorly for making predictions on large domain sizes, prohibiting their use in real-world applications.} This serves as the motivation for the work presented here, and we illustrate our developments by training \textit{\textbf{DiffNet}}, a \emph{data-free} conditional generative model, to solve a parametric family of PDEs. DiffNet belongs to the full-field predictions and PDE family solver classification in the taxonomy introduced above. As such, it serves as a canonical example of a complicated neural architecture that predicts full field outputs for a space of initial/boundary conditions defining a PDE class. We specifically focus on training DiffNet to solve the \textit{inviscid Burgers' equation}, which is a fundamental non-linear PDE with wide applicability in fluid mechanics, gas dynamics and acoustics (i.e. conservation laws with shock formation):
 \begin{equation}
     {\partial u \over \partial t} + u {\partial u \over \partial x} = 0
     \label{eq:Inviscid_Burgers}
 \end{equation}
 We seek to solve this PDE for a one-parameter family of initial conditions defined as 
 \begin{equation}
 u(x,t=0) = \frac{1}{2}(1 - cos(2\pi c x))
 \label{eq:initial-condition}
 \end{equation}
 where $c \ge 0$ is the parameter, and the domain of interest is the unit square, $(x,t) \in \color{black}[0,1]\times[0,0.2]$. Conventional numerical strategies for solving this PDE require some stabilization to gracefully resolve the formation of shocks, and can be computationally expensive for resolved simulations. Fig \ref{fig:fe-m1024-c03} shows a representative solution (generated via space-time finite element solution) for a $1024 \times 1024$ mesh. This took about 400 seconds on 1 SKX node on TACC Stampede2, and serves as our comparative baseline for performance. 


DiffNet is a \textit{convolutional generative} neural network  that takes in instances of parameterized boundary conditions as input and outputs a full field. In order to train DiffNets, we leverage the form of the PDE and minimize the sum of two losses: PDE residual error, and reconstruction error of the initial and boundary conditions. This approach has two major advantages: (1) we only need to \textit{train a single neural network} for the entire parametric family of initial/boundary conditions and/or coefficients, thus allowing fast inference for users; (2) \textit{being data-free}, we do not need any prior solutions of the PDE. 
However, our prior experience with DiffNets~\cite{shah2019encoding} revealed that training DiffNets for larger domain sizes ($> 512 \times 512$) is often impossible on standard GPUs (even on state-of-the-art NVIDIA Tesla V100's). Stable training for such generative models also requires large batch sizes which leads to increasingly larger GPU memory requirements. 

Our primary contribution is a generalized approach to train such large neural network architectures (that are data free) which can serve as (near) real-time neuralPDE solvers. Our main contributions include \textit{(a) a software framework (called \textbf{DeepFusion}) for data parallel distributed deep learning, (b) a hybrid distributed programming approach using OpenMP + MPI for efficient inter/intra node communication, (c) leveraging Intel MKLDNN for very fast forward and back propagation, (d) synchronized batch normalization, (e) loss integrity independent of number of processes, (f) support for Hessian based optimization methods, (g) illustrating this framework to train DiffNet models for $1024\times1024$ domain sizes, which was hitherto not possible on GPUs, and (h) providing results that show nearly $100\times$ speed-up in time to solve a PDE using DiffNets compared to conventional PDE solvers, considering only inference-time.} 

\section{Mathematical preliminaries}
\label{sec:math_prelim}
Using the notation in Hsieh~\etal~\cite{HsiehPDE2019}, we consider a PDE defined as
\begin{eqnarray}
\label{eq:pde}
\gA_\nu(\rvu) = f,~~
\gB(\rvu) = \rvb
\label{eq:bound}
\end{eqnarray}
where $\rvu$ is the solution to the PDE over the domain $\Omega \in \sR^s$, $\gA_\nu$ is the non-linear functional form of the PDE defined by its coefficients $\nu$, and $f$ is a forcing function. Here, $\gB(\cdot)$ refers to the boundary conditions for the PDE. Without loss of generality, we assume that $\Omega$ is the unit square.

\subsection{DiffNets}
\label{subsec:diffnet}

\begin{figure}[htp]
    
    \fbox{\includegraphics[width=0.98\columnwidth]{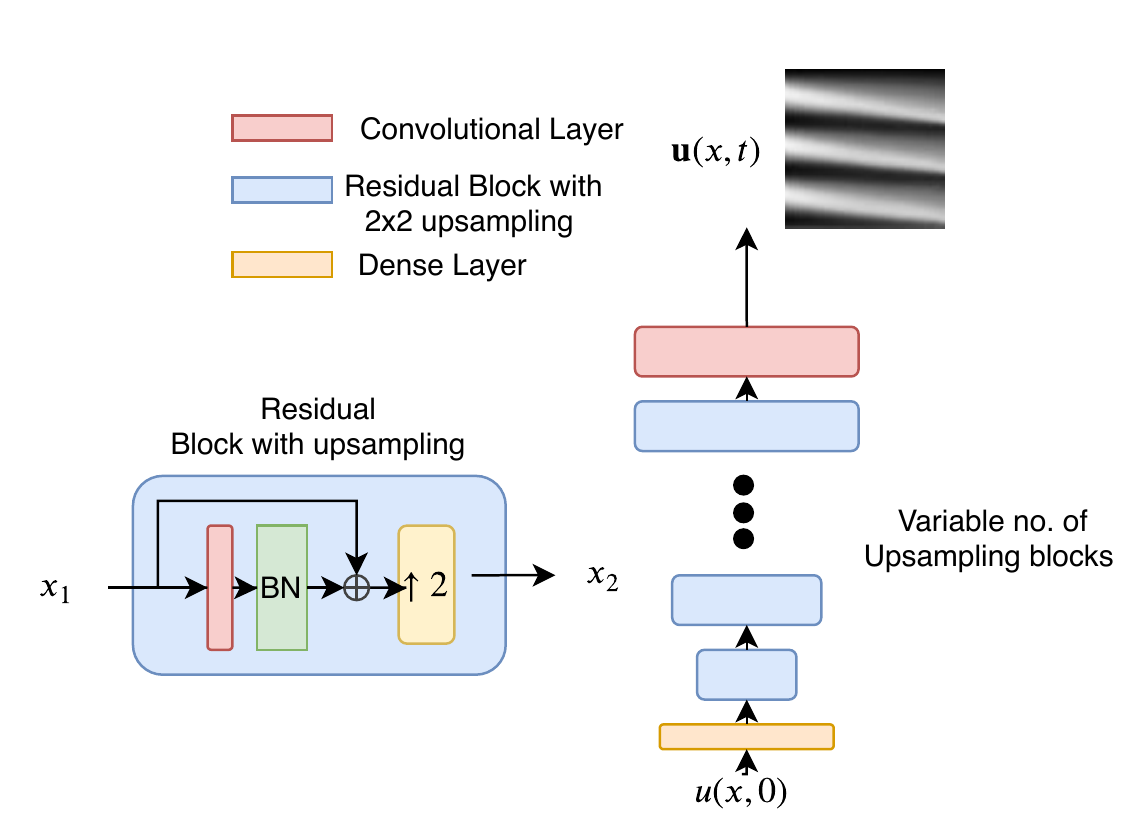}}
    \caption{An exemplar architecture of DiffNets. A specific initial condition, $\rvu(x,0)$ is given as input to the generative model, which then generates the solution for the specified initial value problem. The primary building blocks for the network include residual blocks with upsampling operations. The number of upsampling blocks in the network depends on the resolution of the domain.}
    \label{fig:arch}
\end{figure}

For numerically solving the PDE, the standard approach is to discretize $\Omega$ into ${\rmS} \in \sD^s$ where $\sD$ is a discrete subspace of $\sR^s$. Subsequently, $\rvu$ can be discretized into a vector, $\bar{\rvu}$, by approximating via a basis of piecewise-constant functions over each sufficiently small discrete element. One can similarly also discretize the boundary conditions appropriately. Given a guess solution, $\bar{\rvu}$, standard Finite Difference or Finite Element strategies (FDM, FEM) linearize the non-linear PDE about this guess solution (to get the PDE Jacobian, $\rmA_{\bar{\rvu}}$) and iteratively solve
\begin{eqnarray}
    \rmA_{\bar{\rvu}}(\delta \bar{\rvu}) = \text{res}(\bar{\rvu}); \label{eq:PDE_Solve} \\
    \bar{\rvu} \leftarrow \bar{\rvu} + \delta \bar{\rvu}
\end{eqnarray}
where $\text{res}({\bar{\rvu}})$ is the residual of the current guess w.r.t. the PDE. 
The key computational cost lies in the repeated solution to the linear equation, Eq.~\ref{eq:PDE_Solve}, while computing the residual is computationally trivial. Most modern 'NeuralPDE' approaches exploit this computational asymmetry -- checking to see if a guess $\bar{\rvu}$ is in fact a solution is far more computationally cheaper than actually solving the PDE for the solution, since computing the residual is cheap. 

The DiffNet approach is built on this concept. We model the solution space using a generative neural network. DiffNet consists of a generator $G_\theta:\sR^k\to\sR^d$ that takes as input the initial/boundary conditions $\rvb$ and any PDE coefficients $\nu$. The generator is then trained to generate the solution to the PDE that corresponds to these initial/boundary conditions and coefficients. This also models the stochastic case where $\rvb$ and $\nu$ are sampled from distributions themselves.

We observe that for $G_\theta(\cdot)$ to successfully represent the solution space of the PDE, generator outputs must satisfy two conditions: (1) $G_\theta(\cdot)$ must satisfy the PDE, and (2) $G_\theta(.)$ must respect the provided initial/boundary conditions. The training loss can therefore be written in terms of two components:
\begin{eqnarray}
    & L = L_{p} + \lambda L_{b}, \label{eq:diffnetloss}\\
    \text{where } & L_{p}(\theta) = \E_{\rvb, \nu}[\| \gA_\nu(G_\theta(\rvb, \nu)) - f\|^2_2], \label{eq:pde_loss} \\
    & L_b(\theta) = \E_{\rvb}\|\gB(G_\theta(\rvb, \nu)) - \rvb\|^2_2].
    \label{eq:boundaryloss}
\end{eqnarray}

The first term, $L_p$, minimizes the residual of the PDE while the second term, $L_b$, pushes the generator to learn to reproduce the given initial/boundary conditions. 
\textcolor{black}{As stated in the introduction, this is the overarching strategy for a variety of neural-PDE solvers (e.g., PINN ~\cite{raissi2017physics1} and other works such as ~\cite{zhu2019physics}). The distinction of our approach lies in our choice (a) of predicting the full field, $\rvu(\rvx)$, rather than a single point in the domain. This allows natural enforcement of boundary and initial conditions; and (b) of using a generative model in contrast to other recent approaches. Generative models naturally account for uncertainty, and the network can be extended to produce higher resolution outputs in a straight forward way.}

In order to train the above network, we sample from the space of possible boundary conditions and coefficients, $\{\rvb_i, \nu_i\}, i=\{1, 2, \cdots, k\}$ and optimize the summed loss with respect to $\theta$ using stochastic gradient descent (or a variant such as Adam). Using minibatches sampled from a distribution of $\rvb$ and $\nu$ allows the generator to learn the solutions for the family of PDEs parameterized over $(\rvb, \nu)$.

\noindent \par \textbf{Implementing the forward model.} The derivatives of $L_p(\theta)$ with respect to $\theta$ require calculating $\frac{\partial \gA_\nu}{\partial \theta}$. This is generally non-trivial and to make this tractable we borrow ideas from finite difference methods. We approximate the $k^{th}$ order derivative operator, $\nabla^k_{(\rvx,t)}$ with convolutional operators defined using finite-difference kernels. In practice, we use $3\times3$ Sobel kernels~\cite{sobel19683x3} for first order derivatives and Laplacian kernels~\cite{gonzalez2016dip} for second order derivatives. This is identical to the approach adopted by Zhu~\etal~\cite{zhu2019physics}; however, their setup is somewhat restrictive since they use Encoder-Decoder (ED) networks to construct solutions for a given specific PDE. 

In the case of time-dependent PDEs, the generator $G_\theta$ must learn to first reproduce the initial condition $\rvu_0$ at $t=0$ in order to successfully generate the rest of the solution. An incorrect choice of the Lagrangian coefficient, $\lambda$, leads to failure either by the model learning to generate the trivial solution ($0$) or failing to converge. Additionally, the derivative operators in $x$ and $t$ need to be scaled appropriately in order to satisfy the {\color{black}Courant–Friedrichs–Lewy} condition for stability.  

We show an exemplar architecture for a DiffNet in \Figref{fig:arch}. Note that we rely on additional residual upsampling blocks for finer resolution, so as to keep the parameter count low. The advantage of training a conditional generative model such as DiffNet is that we only need to train a \emph{single} model for a distribution of parameters characterizing the system. Our approach allows for interpolating and (possibly) extrapolating over unseen boundary conditions and coefficients to generate solutions. While our approach uses convolutional layers to reduce the number of parameters, standard GPU based training still restricts us to solving PDEs for limited domain sizes. However, scaling our method to large scale distributed training allows us to bypass this specific disadvantage. In the following section, we discuss our approach for distributed training of DiffNets.

\section{Algorithmic developments}

GPUs remain the overwhelmingly popular compute platform for training these models. GPU memory utilization during training is driven by three factors: 1) number of model parameters in the network, 2) mini-batch size, and 3) size of intermediate tensors created during loss and gradient computations. A known limitation of GPUs is their relatively small available memory: for example, a state-of-the-art NVIDIA Tesla V100 GPU has only 32GB memory. Peak memory utilization to train a DiffNet for domain size $512 \times 512$ and mini-batch size 64 is $\sim$64GB, which is twice the available GPU memory. Due to these memory limitations, training on GPUs is done using mini-batches as small as 16, which in turn results in slow convergence and prohibitive wall-clock times. On Table \ref{tab:gpu_runtimes_memory}, we show maximum batch size and GPU memory utilization for different domain sizes that we were able to train on a NVIDIA Tesla RTX with 24GB memory. \textit{DiffNet training on domain sizes $> 512 \times 512$ is not feasible on currently available GPUs including Tesla V100}.
 \begin{table}[!ht]
  \begin{center}
    \begin{tabular}{|c|c|c|c|}
    \hline
     {\bf Domain Size} &  {\bf Batch Size} & {\bf GPU Memory (GB)} & {\bf Time/Epoch (s)}\\
     \hline
     128$\times$128 & 16	& 0.7 & 105\\
     \hline
     256$\times$256	& 16	& 2.1 & 340\\
     \hline
     512$\times$512	& 16	& 16.4 & 1401\\
     \hline
\end{tabular}
\caption{GPU  memory utilization and time per epoch for 4096 samples for different domain and batch sizes. On Titan RTX with 24GB memory, Diffnet training on domain sizes $> 512 \times 512$ with batch size 16 is not feasible. \label{tab:gpu_runtimes_memory}}
\end{center}
\end{table}

NVIDIA AI Servers like DGX-2 can accommodate bigger batch sizes by distributing the batches across multiple GPUs in a single unit with more cumulative GPU memory (256GB with 8 GPUs and 32GB/GPU). However, they come with an expensive price tag of $\sim$\$0.5M and are not affordable for the general practitioner. In spite of this price tag, the maximum available memory \textbf{is still the same} as the cumulative memory available on 1 or 2 nodes of a CPU cluster.
 
In order to overcome those memory limitations, our \emph{DeepFusion} framework is based on \emph{data parallel} distributed training on multi-node CPU clusters with 5-10x more memory-per-node than a single GPU, and multiple cores-per-node connected via high-end interconnects with low latency and high bandwidth, which can match or exceed the performance of single GPU. \textcolor{black}{In addition to \textit{data parallelism}, extension to \textit{model parallelism} can further push the envelope on accessible network sizes.} 

The rest of this section is organized as follows: in section \ref{subsec:data_parallel_ddl} we provide details on data parallel training; in section \ref{subsec:sync_bn} we discuss the need for synchronized batch normalization; the OpenMP and MPI based hybrid distribution model is explained in section \ref{subsec:hybrid_distribution_model}; time complexities for computation and communication are discussed in section \ref{subsec:complexities}, and a comparison to open-source software is made in section \ref{subsec:comparison_to_oss}. In Table \ref{tab:notations}, we summarize the notations used in this section. 

\begin{table}[!ht]
  \begin{center}
    \begin{tabular}{|cc|}
    \hline
     $N_s$ & Total number of samples \\
     $b_s$ & Number of samples in a mini-batch \\
     $N_s^{loc}$ & Local number of samples \\
     $b_s^{loc}$ & Local mini-batch size \\
     $N_b$ & Number of mini-batches \\
     $N_w$ & Number of weights in the model \\
     $p$   & Number of MPI tasks in \emph{comm} \\
     $N_t$ & Number of threads per MPI task \\
     $F$   & Forward propagation complexity \\
     $B$   & Backward propagation complexity \\
     $L$   & Loss function \\
     $\theta$ & Model parameters \\
     $G_\theta$ & Gradient w.r.t model parameters \\
     \hline
\end{tabular}
\caption{Notations used in this section \label{tab:notations}}
\end{center}
\end{table}

\subsection{Data Parallel Distributed Deep Learning}
\label{subsec:data_parallel_ddl}
\begin{figure}[!ht]
  \centering
  \includegraphics[width=0.48\textwidth]{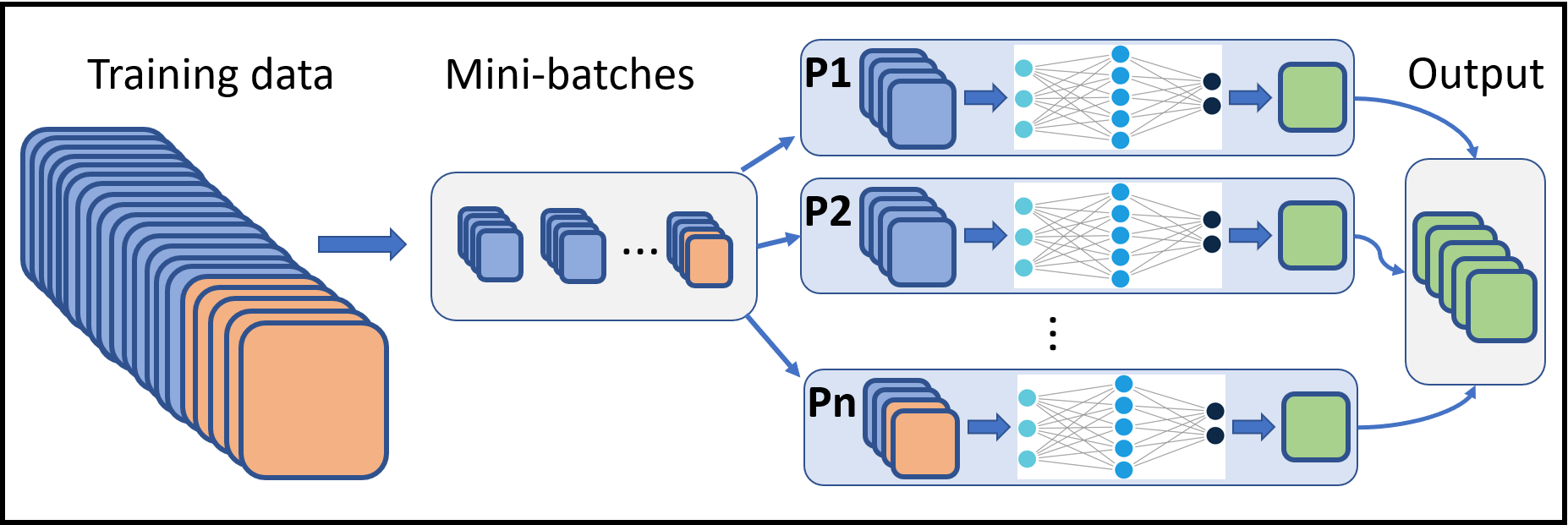}
  \caption{Data Parallel Deep-Learning : multiple replicas of the model are asynchronously trained by workers, each worker using a subset of the global mini-batch. \label{fig:dl_parallel}}
\end{figure}

We use the {\it data parallel} strategy, where multiple replicas of a model are simultaneously trained to optimize a single objective function \cite{bennun_parallel_dnn}. In this approach, the training mini-batches are equally split among the available workers, as shown in Figure \ref{fig:dl_parallel}. Each of these workers asynchronously perform forward and back-propagation of their local mini-batch through the neural network. After each mini-batch, the locally computed gradients are averaged among workers via an {\it MPI\_Allreduce} operation, and that average is used by each local optimizer to update the layer parameters. The loss function is also computed locally, and the objective value is averaged among workers.

\begin{figure}[!ht]
  \centering
  \includegraphics[width=0.48\textwidth]{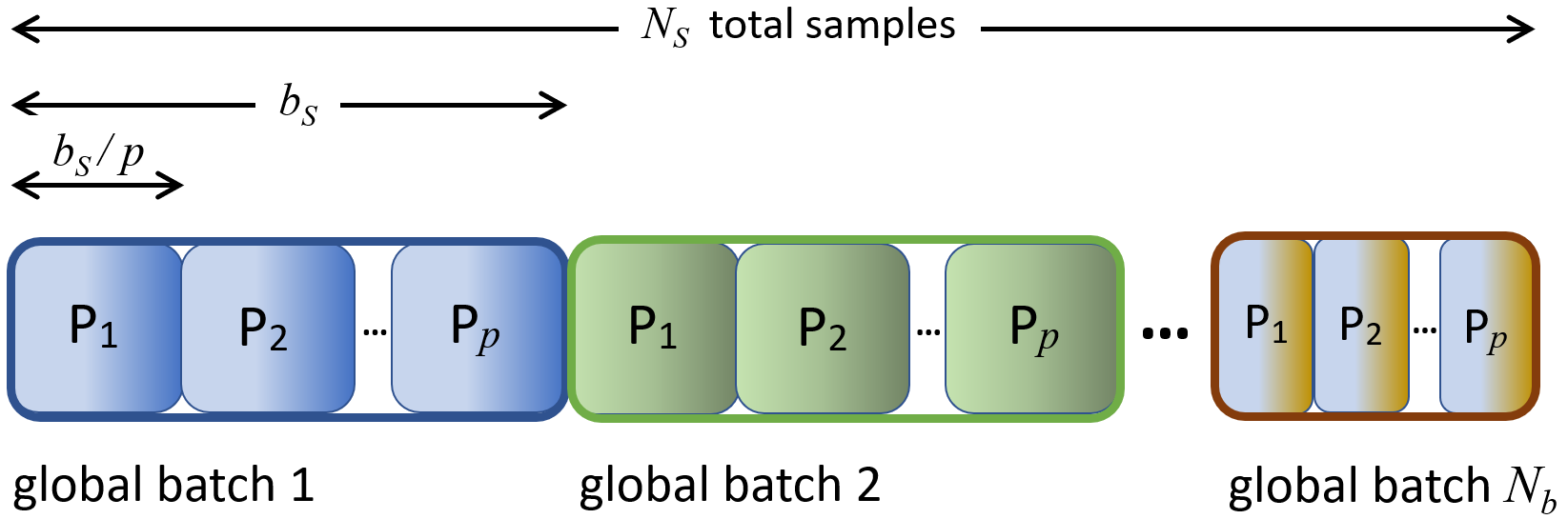}
  \caption{Data splitting across workers in a parallel run: mini-batches are always guaranteed to have the same size across workers at any given time, promoting optimal load balance.\label{fig:mini_batch_split}}
\end{figure}

It is important that the samples (training examples) are split among workers in such a way that the same exact problem gets solved no matter the number of processes (MPI ranks). For that purpose, we adjust the total number of samples $N_s$ and the batch size $b_s$ to make them divisible by the number of workers $p$, such that the {\it local} sample count and batch size are given by
\begin{eqnarray}
N_s^{loc} = \ceil{\frac{N_s}{p}}, \\
b_s^{loc} = \floor{\frac{b_s}{p}}.
\end{eqnarray}
Furthermore, as shown in Figure \ref{fig:mini_batch_split}, each worker draws their local mini-batch sequentially from the global sample pool, in such a way that their union ({\it global} mini-batch) would equal the one used in a single-processor run with the same values of $N_s$ and $b_s$. This guarantees exactly identical results (modulo rounding errors due to gradient reduction) for any number of workers used. Finally, one can easily show that, when $N_s$ is not divisible by $b_s$, the remainder $N_s\mod b_s$ will still be divisible by $p$, which enforces optimal load balancing by guaranteeing that the local mini-batches processed by each worker at any given time have identical sizes.

Algorithm \ref{alg:dl_parallel_algo} summarizes our per-epoch strategy for distributed training using the data parallel paradigm.

\begin{algorithm}[H]
\small
\begin{algorithmic}[1]
\Require Generate $p N_s^{loc}$ samples, split in mini-batches of size $b_s^{loc}$
\For{$mb = 1 \textbf{ to } N_b$}
  \State ${\mathbf{x}}_i \gets \{b_s^{loc} \text{ samples}\}$                  \Comment{Worker $i$ local mini-batch}
  \State $\ell_i(\theta) \gets L({\bf x}_i, G_\theta(\rvb, \nu))$   \Comment{Forward pass and local loss}
  \State ${\bf g}_i(\theta) \gets \nabla\ell_i(\theta)$             \Comment{Back-prop and local gradient}
  \State $\ell(\theta) \gets \frac{1}{p} \sum \ell_i(\theta)$       \Comment{Average loss using MPI\_Allreduce}
  \State ${\bf g}(\theta) \gets \frac{1}{p} \sum {\bf g}_i(\theta)$ \Comment{Average grad using MPI\_Allreduce}
  \State $\Delta\theta \gets u({\bf g}, \theta, t)$                 \Comment{Run local optimizer}
  \State $\theta \gets \theta + \Delta\theta$                       \Comment{Update network parameters}
  \EndFor
\end{algorithmic}
\caption{Data parallel distributed deep learning (1 epoch)\label{alg:dl_parallel_algo}}
\end{algorithm}

    
    

\subsection{Synchronized Batch-normalization}
\label{subsec:sync_bn}
Batch Normalization (BN) is a procedure that dramatically improves the convergence of neural networks by re-scaling and re-centering data using running statistics, namely mean and variance, accumulated from each mini-batch in the course of an epoch \cite{ioffe_batchnorm}. This creates a dependency on the local mini-batch size, breaking the paradigm of problem independence on the number of workers discussed in section \ref{subsec:data_parallel_ddl}.
To remove this dependency, we developed a scheme to synchronize the mean and variance statistics at all BN layers by performing an {\it MPI\_Allreduce} operation after each epoch. This is especially important when the local mini-batch size on each processor is small, which would result in poor statistics and have a negative impact on validation accuracy. The BN synchronization scheme is explained in Algorithm \ref{alg:dl_bn_sync}.

\begin{algorithm}[H]
\small
\begin{algorithmic}[1]
\For{\textbf{each} ${epoch}$} 
\For{\textbf{each} $BN$ $layer$ $in$ $network$} 
\If{$in$ $evaluation$} 
   \State $\mu_{\mathcal{B}} \gets \frac{1}{p} \sum_i^p \mu_{\mathcal{B}}^{(i)}$ \Comment{Allreduce BN means}
   \State $\sigma_{\mathcal{B}}^2 \gets \frac{1}{p} \sum_i^p {\sigma_{\mathcal{B}}^2}^{(i)}$ \Comment{Allreduce BN variances}
\EndIf
\EndFor
\EndFor
\end{algorithmic}
\caption{Batch-normalization synchronization algorithm\label{alg:dl_bn_sync}}
\end{algorithm}

\subsection{Hybrid Distribution Model}
\label{subsec:hybrid_distribution_model}
\begin{figure}[!ht]
  \centering
  \includegraphics[width=0.45\textwidth]{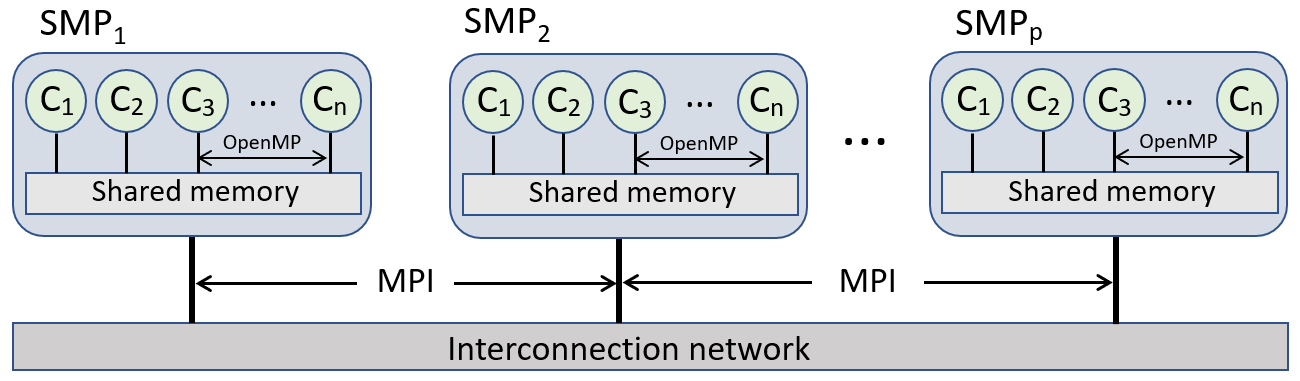}
  \caption{Process-to-process hybrid distribution paradigm: processes communicate via MPI, and spawn local threads of their own that communicate via OpenMP inside an SMP node. MPI routines are only invoked outside of OpenMP parallel regions.  \label{fig:hybrid_distribution}}
\end{figure}

Our parallel distribution scheme is based on the so called {\it hybrid} MPI-OpenMP programming paradigm, in which communication between processes is done via MPI, while each process can spawn its own OpenMP threads that run inside a single shared-memory processor (SMP) node, as illustrated in Figure \ref{fig:hybrid_distribution}. Furthermore, since the OpenMP threads only communicate with other threads within the same SMP node, and MPI routines are only invoked outside of OpenMP parallel regions, our distribution scheme can be said to model the {\it process-to-process} hybrid paradigm.
In particular, our application spawns $p$ processes via the usual \texttt{mpirun} utility, which can land on up to $p$ SMP nodes. The number of processes per node depends on the specific specs of the host machines and on details of the experiment. A few underlying libraries used by our application (e.g., libtorch and mkldnn) spawn up to $N_t$ local threads of their own, where $N_t$ can be controlled via the OMP\_NUM\_THREADS environment variable. 

\subsection{Complexities}
\label{subsec:complexities}
In this sub-section, we will discuss computation and communication complexities of our approach. As shown in Figure \ref{fig:dl_parallel}, the entire data set with $N_s$ samples is split into $N_b$ mini-batches with $b_s$ samples per mini-batch. Each MPI task computes loss and gradients for the local mini-batch of size $b_s^{loc}$. Loss and gradient computations have a complexity of 
\begin{equation}
O\left(F(N_w,b_s^{loc}) + B(N_w, b_s^{loc})\right).
\end{equation}
Forward ($F$) and backward ($B$) propagation complexities are non-linear functions of $N_w$, $b_s^{loc}$, number of cores allocated per MPI task and the network architecture. We use libtorch C++ APIs to execute forward and backward propagation in a single MPI task, which internally uses MKLDNN \cite{mkldnn} for optimal performance on Intel CPUs. 
Local gradients are averaged using  \emph{MPI\_Allreduce}, which has a communication complexity of $O(N_w + log(p))$. Since $N_w\gg  p$, we expect the communication time to remain almost constant, independent of $p$ and of the underlying algorithm used by the OpenMPI implementation. Network weights on each process are updated using this averaged gradient via a stochastic gradient descent (SGD) method. Synchronization of statistics for all batch normalization layers after each epoch has a complexity of  $O(N_w)$. The overall complexity per epoch is, therefore,
\begin{equation}
    O\left(N_b\left(F(N_w,b_s^{loc})+ B(N_w,b_s^{loc}) + N_w \right) \right).
\end{equation}

\subsection{Comparison to Open Source Software}
\label{subsec:comparison_to_oss}
\textit{\textbf{Tensorflow}} and \textit{\textbf{Pytorch}} are the most popular open source libraries for deep learning, and both provide support for distributed deep learning. Out of the box, they handle all-reduce operations on gradients computed on a local mini-batch on each device. \textit{\textbf{Horovod}} is an open source library \cite{sergeev_horovod} from Uber that enables data parallelism and gradient averaging. However, optimal load balancing, guarantees of loss independence on processor count, $p$ (Figure \ref{fig:loss_vs_epoch_256x256}), synchronized batch normalization and hybrid distributed implementation discussed in the previous sections are not provided out of the box by these open source libraries. 

DeepFusion democratizes the ability of data scientists to train models that are too big for GPUs with desired system performance and convergence rates without any distributed and high performance computing experience.

\begin{table}[!ht]
  \begin{center}
    \begin{tabular}{|c|c|c|c|c|} 
      \hline
      \shortstack{\textbf{Out of the box} \\ \textbf{functionality}} & \textbf{TF} & \textbf{Pytorch} &
       \textbf{Horovod} & \textbf{DeepFusion}\\
      \hline
      \shortstack{All reduce on \\Gradient} & Yes & Yes & Yes & Yes \\
      \hline
      \shortstack{Loss Integrity \\ independent of $p$} & No & No & No & {\bf Yes} \\
      \hline
      \shortstack{Synchronized \\ Batch Normalization} & No & No & No & {\bf Yes} \\
      \hline
    \end{tabular}
    \caption{Qualitative comparison of DeepFusion functionality with Open Source Software \label{tab:oss_comparison}}
  \end{center}
\end{table}


\section{Results and Discussions}
One of the key outcome of our experiments was to demonstrate a practical approach to train DiffNets on domain sizes $> 512 \times 512$ that are too big for GPUs. We tested our framework on both TACC \textbf{Stampede2} HPC clusters with bare-metal access, as well as \textbf{Microsoft Azure} and \textbf{Amazon Web Services} (AWS) HPC clusters built using on-demand virtual machines. We target these computational resources as representative of what is easily accessible for the general data science practitioner unlike DGX-2 that requires significant investment. We report results obtained from \ipoint{AWS}, Microsoft Azure and Stampede2. On Table \ref{tab:aws_azure_vs_stampede_specs}, we provide all relevant specifications for Azure and Stampede2 used in our experiments. \ipoint{Care was taken to select configurations on AWS and Azure to reasonably match the CPU as well as interconnect speeds of Stampede2. This allows rational assessment of performance of DeepFusion across nearly similar platforms.} We present wall-clock time comparisons between \ipoint{AWS}, Azure and Stampede2 in section \ref{subsec:conventional_cloud} to determine the cluster to use for our large domain runs. In section \ref{subsec:strong_scaling}, we conduct strong scaling experiments for $128 \times 128$ and $256 \times 256$ domain sizes for 1 to 128 nodes (48 to 6144 cores) on Stampede2 to demonstrate scalability of our software. Finally, in section \ref{subsec:high_resolution_diffnet}, we present results for training a DiffNet model for \textbf{$512 \times 512$} and \textbf{$1024 \times 1024$} domain sizes for Burgers' inviscid equation (Eq.~\ref{eq:Inviscid_Burgers}) for parameter distributions characterizing the initial conditions (Eq.~\ref{eq:initial-condition}), which are currently not possible to train.

\begin{table}[!ht]
  \begin{center}
    \begin{tabular}{|c|c|c|c|}
    \hline
     {\bf Specification} & \ipoint{\bf AWS} & \shortstack{{\bf Azure}}  & {\bf Stampede2} \\
     \hline
     Type & \ipoint{\shortstack{Virtual \\Machine}} & \shortstack{Virtual \\ Machine} & Bare-Metal \\
     \hline
     CPU & \ipoint{\shortstack{Intel Xeon \\ Platinum 8000}} & \shortstack{Intel Xeon \\ Platinum 8168} & \shortstack{Intel Xeon \\ Platinum 8160} \\
     \hline
     CPU cores & \ipoint{72} & 44 & 48 \\
     \hline
     Memory (GB)& \ipoint{192} & 352& 192\\
     \hline
     Interconnect & \ipoint{\shortstack{Elastic \\ Fabric Adapter}} & \shortstack{EDR \\Infiniband} & \shortstack{Intel \\ Omni-Path} \\
     \hline
     Bandwidth & \ipoint{100 Gb/sec} & 100 Gb/sec & 100 Gb/sec \\
     \hline
     Topology & \ipoint{AWS Proprietary} & Fat tree & Fat tree \\
     \hline
\end{tabular}
\caption{Functional specifications of \ipoint{AWS}, Microsoft Azure and Stampede2 infrastructure used in our experiments. \label{tab:aws_azure_vs_stampede_specs}}
\end{center}
\end{table}

\subsection{Conventional HPC vs. Cloud Based HPC} 
\label{subsec:conventional_cloud}
In this section, we compare wall-clock times between \ipoint{AWS}, Microsoft Azure, and Stampede2 in order to determine the optimal HPC cluster configuration to train \emph{DiffNet} with very large ($1024\times 1024$) resolutions. On Table \ref{tab:aws_azure_vs_stampede}, we show per-epoch wall-clock times (in seconds) to train DiffNet with $64\times 64$ resolution using 1, 2 and 4 nodes. The total number of samples used for this experiment was $N_s = 4096$ and the global batch size was $b_s = 1024$; the number of processes per node was fixed at 4, with each process spawning 8 local threads. Single node performance on bare-metal Stampede2 is $\sim 2 \times$ faster than on Azure \ipoint{and AWS} VM. On the same table, we also compare per-epoch wall-clock times (in seconds) for different resolutions on 4 compute nodes (with 4 processes per node, 8 threads per process). Even though the infrastructure specifications of Azure, \ipoint{AWS} and Stampede2 are almost identical, slowness on Azure and \ipoint{AWS} can be attributed to the overheads associated with virtual machines.

\begin{table}[!ht]
  \begin{center}
    \begin{tabular}{|c|c|c|c|c|}
    \hline
     {\bf Domain Size} & {\bf Nodes} & \ipoint{\bf AWS} & {\bf Azure} & {\bf Stampede2}  \\
     \hline
     64x64 & 1 & \ipoint{131.0} & 113.1&67.2 \\
     \hline
     64x64 & 2 & \ipoint{65.2} & 54.9&34.9 \\
     \hline
     64x64 & 4 & \ipoint{32.4} & 28.6& 19.4\\
     \hline
     128x128 & 4& \ipoint{138.4} & 126.2 &68.5 \\
     \hline
     256x256 & 4 & \ipoint{650.5} & 597.6& 279.8\\
     \hline
\end{tabular}
\caption{Comparison of per-epoch wall-clock times (in seconds) between \ipoint{AWS}, Azure and Stampede2 for varying resolutions and different number of nodes (see Table \ref{tab:aws_azure_vs_stampede_specs} for cluster specs). For all three clusters, 4 processes were used per node (spawning 8 threads each). \label{tab:aws_azure_vs_stampede}}
\end{center}
\end{table}
\subsection{Scaling Experiments} 
\label{subsec:strong_scaling}
\begin{figure}[!ht]
  \centering
  \includegraphics[width=0.45\textwidth]{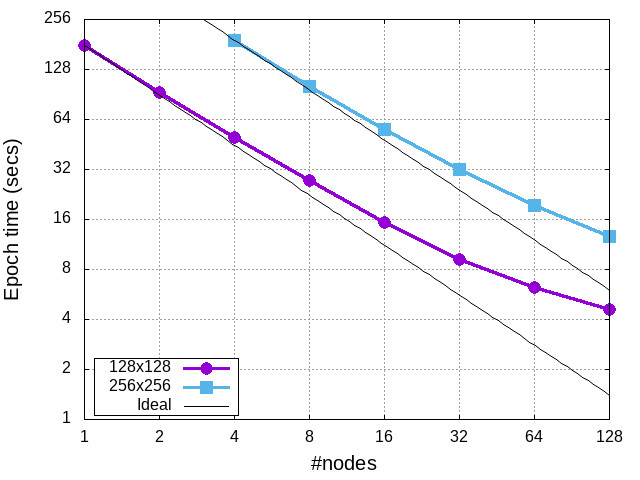}
  \caption{Strong scaling results for training {\it DiffNet} on Stampede2: per-epoch times (in seconds) versus number of nodes for $128\times 128$ and $256\times 256$ resolutions, using 8 processes per node and 12 threads per process.\label{fig:sc_strong_scaling}}
\end{figure}

In Figure \ref{fig:sc_strong_scaling}, we report strong scaling results to train {\it DiffNet} for $128\times 128$ and $256\times 256$ resolutions, using from 4 to 128 nodes on Stampede2. In this experiment, we used 8 MPI processes per node and each process spawned 12 threads, to a total of 96 threads per node. This matches the full capacity of the Stampede2 Skylake nodes, which have 48 physical hyperthread-enabled CPU cores, resulting in 96 hardware threads per node. 
\ipoint{In Table \ref{tab:wall_clock_times}, we compare per-epoch wall clock time between a single GPU (Titan RTX as well as Tesla V100) with the wall clock time using 128 Stampede2 nodes. We show this (potentially unfair) comparison to illustrate the advantage of scale-up on CPUs using a distributed training approach}.

\begin{table}[!ht]
  \begin{center}
    \begin{tabular}{|c|c|c|c|}
    \hline
     \thead{\bf Output \\ \bf resolution} &   \thead{\bf{1 Titan RTX} \\ \bf{(seconds)}} &   \thead{\bf{1 Tesla V100} \\\bf{(seconds)}} & \thead{\bf 128 Stampede2 nodes\\  \bf (seconds)} \\
     \hline
     128$\times$128 & 105 & \ipoint{130} & 4.6\\
     \hline
     256$\times$256	& 340 & \ipoint{494} & 12.7\\
     \hline
     512$\times$512	& 1401 & \ipoint{1961} & 25.3 \\
     \hline
     1024$\times$1024 & N/A & \ipoint{N/A}& 89.5 \\
     \hline
\end{tabular}
\caption{Comparison of per-epoch wall-clock time  between Titan RTX, \ipoint{Tesla V100} and DeepFusion on 128 Stampede2 nodes to train Diffnet (of different resolutions) with 4096 samples. Training on large CPU clusters using DeepFusion is $20-60\times$ faster than training on \ipoint{both GPU's. The 30-40\% change between the Titan RTX vs Tesla V100 is attributable to the 30\% difference in clockspeed between them.}\label{tab:wall_clock_times}}
\end{center}
\end{table}

\begin{figure}[!ht]
  \centering
  \includegraphics[width=0.45\textwidth]{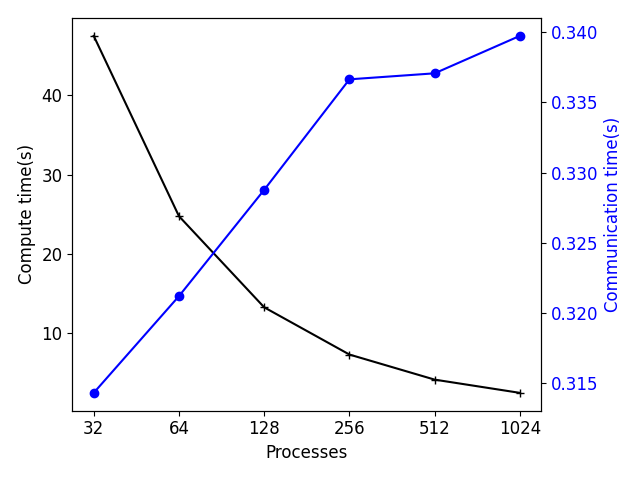}
  \caption{\ipoint{Computation and communication wall clock times for different $p$ corresponding to strong scaling results for domain size $256 \times 256$ in Figure \ref{fig:sc_strong_scaling}. Computation time decreases linearly with $p$, while communication time increases only slightly with $p$.} \label{fig:comp_vs_comm_wc_times_256x256_revised}}
\end{figure}

As discussed in section \ref{subsec:complexities}, our computation time complexity scales with $p$ and communication complexity is independent of $p$. In Figure \ref{fig:comp_vs_comm_wc_times_256x256_revised}, we show the computation and communication wall-clock times for different $p$. We observed that our communication times \ipoint{increase only slightly with $p$, as expected. Note that the communication times are significantly ($100\times$) smaller than compute times}.

\begin{figure}[!ht]
  \centering
  \includegraphics[width=0.45\textwidth]{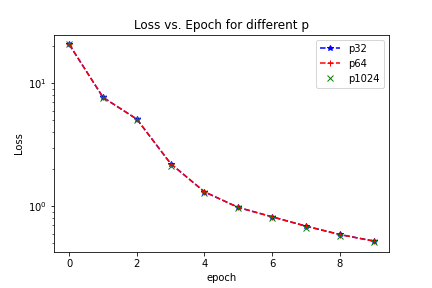}
  \caption{Comparison of loss decay vs. epoch for a subset of $p$ used in strong scaling experiments in Figure \ref{fig:sc_strong_scaling}. Loss decay is preserved (independent of $p$). \label{fig:loss_vs_epoch_256x256}}
\end{figure}

As discussed in section \ref{subsec:data_parallel_ddl}, the training samples are split in such a way as to guarantee loss decay integrity, i.e., the same exact problem is solved independently of the number of MPI processes. In Figure \ref{fig:loss_vs_epoch_256x256}, we show the loss vs. epoch for different values of $p$. The small deviation in loss values for different $p$ is due to rounding errors in \emph{MPI\_Allreduce} operations for computing gradient averages.

\begin{figure*}[ht]
\centering
\captionsetup{justification=centering}
    \begin{subfigure}{0.4\textwidth}
      \centering
      \subcaption*{DeepFusion solution}
      \includegraphics[trim=80 50 80 80,clip,width=1.0\linewidth]{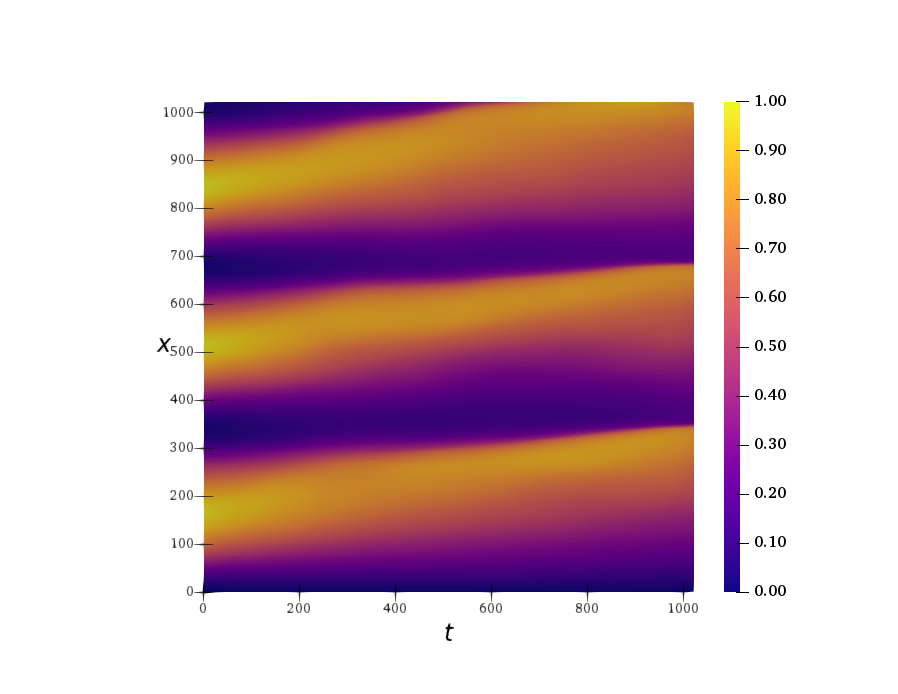}
    \end{subfigure}
    \begin{subfigure}{0.4\textwidth}
      \centering
      \subcaption*{Finite element solution}
      \includegraphics[trim=80 50 80 80,clip,width=1.0\linewidth]{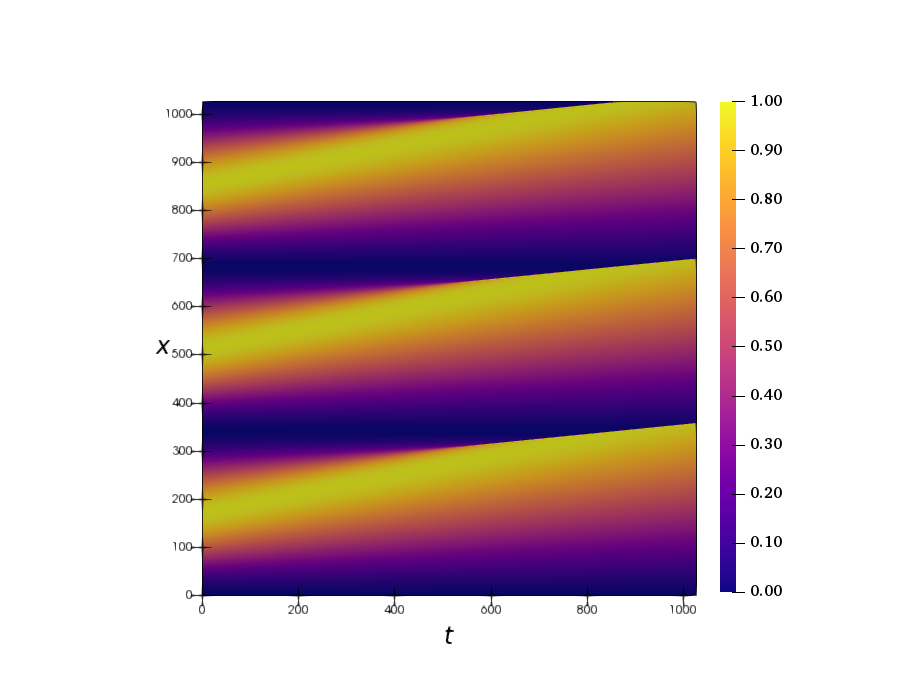}
    \end{subfigure}
    \caption{Inviscid Burgers' equation  solved on $1024\times1024$ pixels (left) or elements (right)}
    \label{fig:comparison-panel_1024}
\end{figure*}

\begin{figure}[h]
  \centering
  \includegraphics[width=0.45\textwidth]{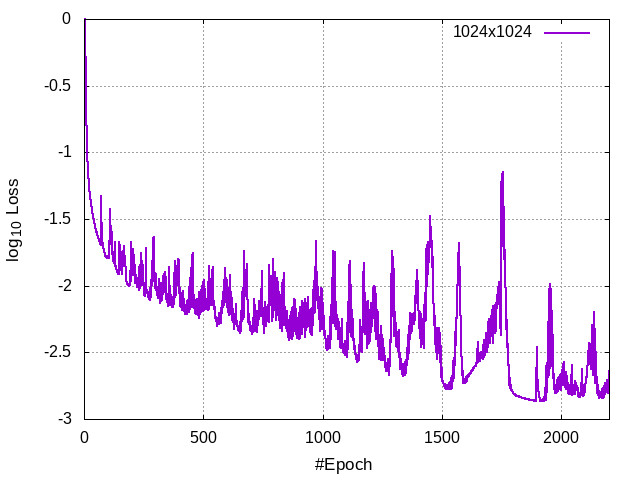}
  \caption{Loss vs. epoch for training DiffNet with domain size $1024\times 1024$ on Stampede2 using 256 sample points and batch size 64.\label{fig:loss_1024}} \vspace{-0.2in}
\end{figure}

\begin{figure}[h]
  \centering
  \includegraphics[width=0.45\textwidth]{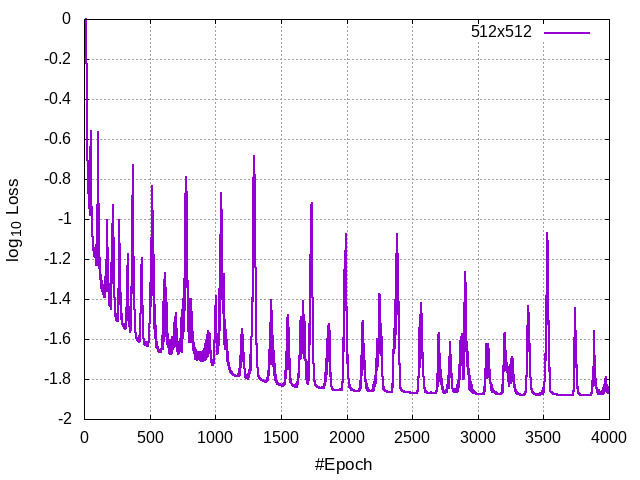}
  \caption{Loss vs. epoch for training DiffNet with domain size $512\times 512$ on Stampede2 using 256 sample points and batch size 64.\label{fig:loss_512}}\vspace{-0.2in}
\end{figure}

\begin{figure}[!ht]
    \centering
    \begin{minipage}[t]{.4\textwidth}
      \centering
      \begin{tikzpicture}
      \begin{axis}[
          width=0.99\linewidth, 
          xlabel=$x$, 
          ylabel=$u$,
          legend cell align={left},
          legend style={at={(1,0)},anchor=south east,legend columns=1}, 
          grid=both,
          x tick label style={rotate=0,anchor=north} 
        ]
         \addplot+ [color=black,mark=none,line width=1.2pt]
        table[x expr={\thisrow{"arc_length"}},y expr={\thisrow{"u_ml"}},col sep=comma]{n_512_c_10.csv};
        \addplot+ [color=red,mark=none]
        table[x expr={\thisrow{"arc_length"}},y expr={\thisrow{"u_fem"}},col sep=comma]{n_512_c_10.csv};
        \addplot+ [color=blue,dashed,mark=none,line width=1pt]
        table[x expr={\thisrow{"arc_length"}},y expr={\thisrow{"u_fdm"}},col sep=comma]{n_512_c_10.csv};
        \legend{\scriptsize{DeepFusion}, \scriptsize{FEM, space-time},\scriptsize{FDM, time-marching}}
      \end{axis}
    \end{tikzpicture}
    \end{minipage}
    \caption{\ipoint{An example of the solution profile $u$ vs. $x$ at the final time ($t = 0.2$), compared for different methods. Resolution = $512\times 512$ and $c = 10$}}
    \label{fig:plot-solutions-vs-x-t-1-resolution-512-c-10}
\end{figure}
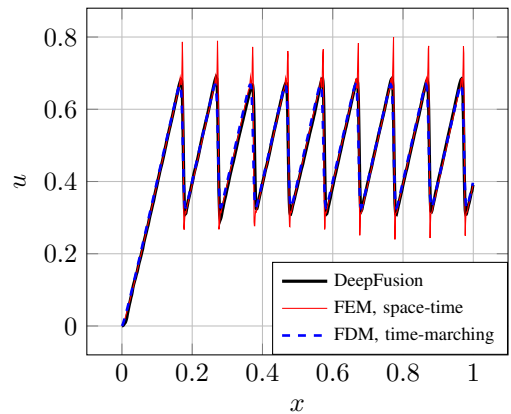

\subsection{High Resolution DiffNet} 
\label{subsec:high_resolution_diffnet}
In this sub-section, we illustrate the ability of the framework to train {\it DiffNet} models with very high resolution outputs (sizes $\ge 512\times 512$). This typically requires $\ge 2000$ epochs until convergence (for different ranges of the initial condition frequency $c$) \ipoint{using a first-order optimizer like SGD}.

{\bf Example 1:} In the first example, we train a DiffNet to produce outputs of resolution $1024 \times 1024$. We emphasize that generative models of this size have hitherto fore not been trained (to the best of our knowledge). We train the DiffNet to be predictive in a range of the parameter $c\in [3,6]$. We remind the reader that $c$ represents a one-parameter family of initial conditions to the inviscid Burgers equation. The training sample set consists of 256 points from $c\in [3,6]$. We set the mini-batch size to 64. This model was trained on 8 nodes of Stampede2 (with 8 processes per node), taking 2200 epochs till convergence (see Fig.~\ref{fig:loss_1024}). The runtime for this training was 32 hours. 

After training, we performed inference using the DiffNet for a set of initial conditions from the one-parameter family. Results for $c = 3$ are shown in Fig.~\ref{fig:comparison-panel_1024}, where we compare the DiffNet inference result with the solution from an optimized FEM solver (based on the Petsc library) of the inviscid Burgers equation. While the general trend of the solution (as the wave evolves -- from left to right in the figures -- and forms shocks) is captured well, there is still room for improvement as the more diffused nature of the ML solution indicates. We hypothesize that using higher order Sobel filters (i.e. computing the gradients using higher order stencils) could help in eliminating the diffusive features of the ML solution. We continue to explore these aspects.




{\bf Example 2:} In the second example, we explore if the DiffNet can be trained to predict solutions for a much larger distribution of the initial conditions. We train a DiffNet to predict solutions at $512\times 512$ resolution, but for initial conditions from $c\in [3,16]$. At higher values of c (which represent initial conditions exhibiting higher frequencies), we expect the formation of multiple shocks. The traditional numerical solutions for these initial conditions have to be carefully performed. We set the mini-batch size to 64. This model was trained on 8 nodes (64 processors) on Stampede2, taking 4000 epochs till convergence (see Fig.~\ref{fig:loss_512}). The runtime for training this model was 15 hours. Notice that the loss in this case is significantly larger than the previous example. This is due to two reasons: the reduced resolution ($1024 \rightarrow 512$) and, more importantly, the larger $c$ space.  

\begin{figure*}[!ht]
\centering
\captionsetup{justification=centering}
    \begin{subfigure}{0.4\textwidth}
      \centering
      \subcaption*{DeepFusion solution}
      \includegraphics[trim=30 0 0 30,clip,width=1.0\linewidth]{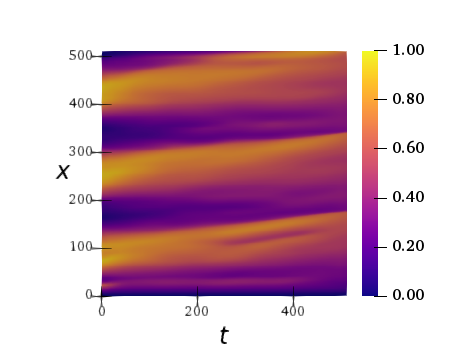}
    \end{subfigure}
    \begin{subfigure}{0.4\textwidth}
      \centering
      \subcaption*{Finite element solution}
      \includegraphics[trim=30 0 0 30,clip,width=1.0\linewidth]{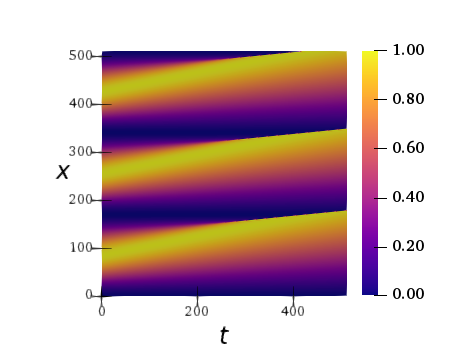}
    \end{subfigure}
    \\
    \begin{subfigure}{.4\textwidth}
      \centering
      \includegraphics[trim=30 0 0 30,clip,width=1.0\linewidth]{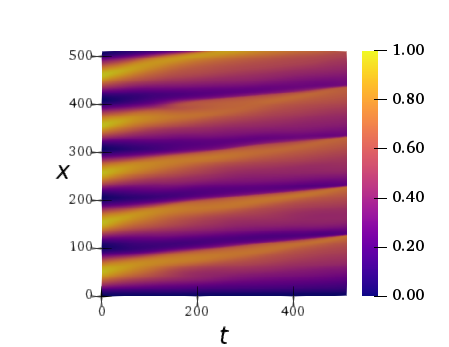}
    \end{subfigure}
    \begin{subfigure}{.4\textwidth}
      \centering
      \includegraphics[trim=30 0 0 30,clip,width=1.0\linewidth]{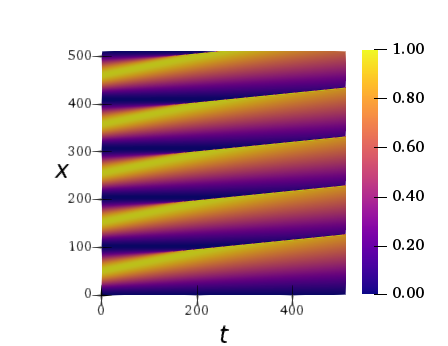}
    \end{subfigure}
    \\
    \begin{subfigure}{.4\textwidth}
      \centering
      \includegraphics[trim=30 0 0 30,clip,width=1.0\linewidth]{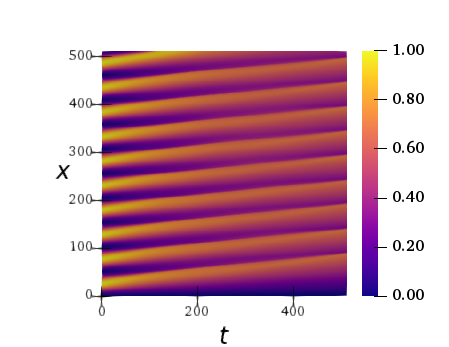}
    \end{subfigure}
    \begin{subfigure}{.4\textwidth}
      \centering
      \includegraphics[trim=30 0 0 30,clip,width=1.0\linewidth]{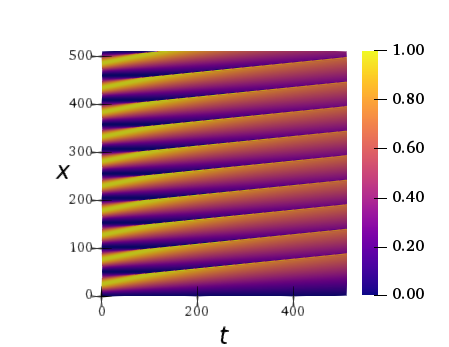}
    \end{subfigure}
    \\
    \begin{subfigure}{.4\textwidth}
      \centering
      \includegraphics[trim=30 0 0 30,clip,width=1.0\linewidth]{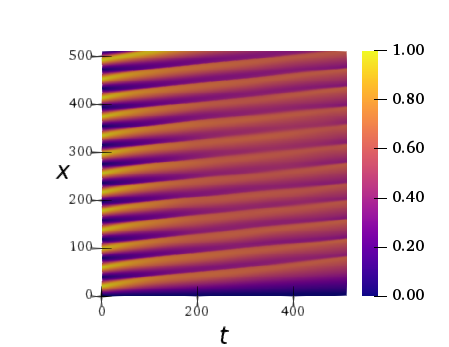}
    \end{subfigure}
    \begin{subfigure}{.4\textwidth}
      \centering
      \includegraphics[trim=30 0 0 30,clip,width=1.0\linewidth]{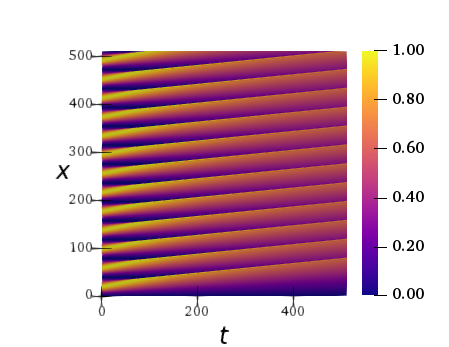}
    \end{subfigure}
    \caption{Contour plots of the solution to the inviscid Burgers' equation: solutions from DiffNet (left) vs solutions from a finite element solver (right). The initial condition is characterized by the wave number $c$ (defined in \ref{eq:initial-condition}). The first row presents solution for $c = 3$ and subsequently for $c$ = 5, 10 and 13 in the latter rows respectively. On the left, the image size is $512\times512$ pixels, whereas on the right, the discretization is a mesh of $512\times512$ bilinear quadrilateral elements.}
    \label{fig:comparison-panel}
\end{figure*}

After training, we again performed inference using the DiffNet for a set of initial conditions from the one-parameter family. Results for $c = 3, 5, 10, 13$ are shown in Fig.~\ref{fig:comparison-panel}, where we compare the DiffNet inference result with the solution from an optimized FEM solver (based on the Petsc library) of the inviscid Burgers equation. As before, the general trend of the solution (as the wave evolves -- from left to right in the figures -- and forms shocks) is captured well, but there is still room for improvement. 

\ipoint{Fig.~\ref{fig:plot-solutions-vs-x-t-1-resolution-512-c-10} and Table~\ref{tab:norms-solutions-and-differences} show additional, quantitative comparison  between the DiffNet results with a stabilized Finite Element solution (at $512 \times 512$ resolution) against a very high resolution ($2048 \times 2048$) finite difference solution. Fig.~\ref{fig:plot-solutions-vs-x-t-1-resolution-512-c-10} plots the solution at one time point ($t=0.2$), and suggests that stabilized finite element space-time approach is still unable to capture the shocks, while the DiffNet is able to accurately capture the shock without any dispersive effects. This is particularly promising as the loss function used in the DiffNet is the simplest one possible, with significant room for improvement. Table~\ref{tab:norms-solutions-and-differences} shows the L2 error norm (in space-time) of the DiffNet and FEM solution against the high resolution FDM solution. Interestingly, we find that the DiffNet approach produces more accurate results for initial conditions exhibiting more waves (larger c). This is in contrast to what is observed in traditional approaches (compare the last two columns of Table~\ref{tab:norms-solutions-and-differences}).} We find it promising that the DiffNet is able act as a general PDE solver for a wide range of initial conditions. This strongly suggests that, with the proper training infrastructure, it is possible to develop truly general PDE solvers that produce accurate solutions for general classes of PDE's. 

\begin{table}[]
    \centering
\csvreader[
  tabular=|c|c|c|c|c|c|c|,
  table head=\hline \bfseries{\boldmath${N}$} & \bfseries{\boldmath${c}$} & \bfseries{\boldmath${||u_g||_2}$} & \bfseries{\boldmath${||u_{fd}||_2}$} & \bfseries{\boldmath${||u_{fe}||_2}$} & \bfseries{\boldmath${||\delta^{g}_{fd}||_2}$} & \bfseries{\boldmath${||\delta^{fe}_{fd}||_2}$} \\ \hline,
  late after last line=\\ \hline 
]{
  norms_table.csv
}{}{\csvlinetotablerow}
    \caption{\ipoint{Norm of different solutions (denoted $u$) and their differences (denoted $\delta$). $u_g$ is the solution generated through DeepFusion, $u_{fd}$ is the solution obtained using explicit time marching with finite difference approximation; and $u_{fe}$ is the ``space-time" solution  obtained through finite element approximation in \textit{both} space and time. The differences between them: $\delta^{g}_{fd} = u_g - u_{fd}$ and $\delta^{fe}_{fd} = u_{fe} - u_{fd}$. All norms are calculated over the entire spatiotemporal domain}}
    \label{tab:norms-solutions-and-differences} 
\end{table}

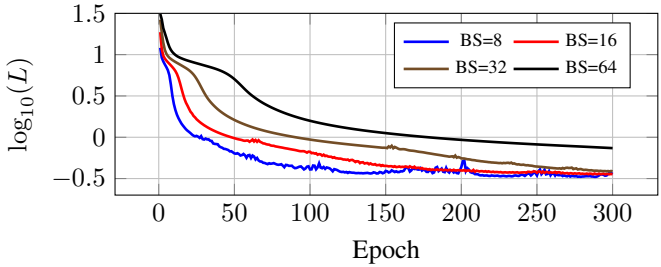
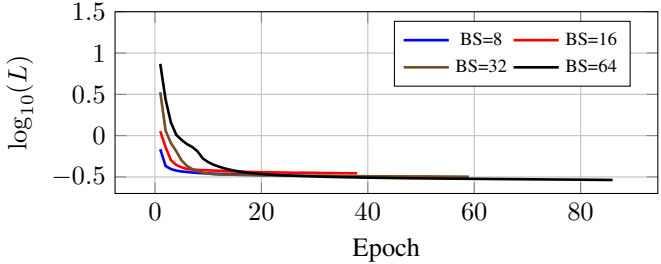
\begin{figure}[!ht]
    \centering
    \begin{minipage}{.49\textwidth}
  \begin{center}
  \begin{tikzpicture}
      \begin{axis}[
          width=0.99\linewidth, 
          height=4cm,
          xlabel={Epoch}, 
          ylabel=$\log_{10}(L)$, 
          legend style={at={(0.95,0.95)},anchor=north east,legend columns=2}, 
          ymin=-0.7,
          ymax=1.5,
          grid=both,
          x tick label style={rotate=0,anchor=north}
        ]
        \addplot+ [mark=none,line width=1pt]
        table[x expr={(\thisrowno{1})},y expr={log10(\thisrowno{3})},col sep=space]{batch-sz-exp-0064-sgd-log_processed_bs_8.txt};
        \addplot+ [mark=none,line width=1pt]
        table[x expr={(\thisrowno{1})},y expr={log10(\thisrowno{3})},col sep=space]{batch-sz-exp-0064-sgd-log_processed_bs_16.txt};
        \addplot+ [mark=none,line width=1pt]
        table[x expr={(\thisrowno{1})},y expr={log10(\thisrowno{3})},col sep=space]{batch-sz-exp-0064-sgd-log_processed_bs_32.txt};
        \addplot+ [mark=none,line width=1pt]
        table[x expr={(\thisrowno{1})},y expr={log10(\thisrowno{3})},col sep=space]{batch-sz-exp-0064-sgd-log_processed_bs_64.txt};
        \legend{\scriptsize{BS=8}, \scriptsize{BS=16}, \scriptsize{BS=32}, \scriptsize{BS=64}}
      \end{axis}
    \end{tikzpicture}
    \subcaption{SGD}
    \label{fig:64x64-sgd-loss-epoch}
    \end{center}
    \end{minipage}
    \\
    \begin{minipage}{.49\textwidth}
  \begin{center}
  \begin{tikzpicture}
      \begin{axis}[
          width=0.99\linewidth, 
          height=4cm,
          xlabel={Epoch}, 
          ylabel=$\log_{10}(L)$, 
          legend style={at={(0.95,0.95)},anchor=north east,legend columns=2}, 
          ymin=-0.7,
          ymax=1.5,
          grid=both,
          x tick label style={rotate=0,anchor=north}
        ]
        \addplot+ [mark=none,line width=1pt]
        table[x expr={(\thisrowno{1})},y expr={log10(\thisrowno{3})},col sep=space]{batch-sz-exp-0064-lbfgs-log_processed_bs_8.txt};
        \addplot+ [mark=none,line width=1pt]
        table[x expr={(\thisrowno{1})},y expr={log10(\thisrowno{3})},col sep=space]{batch-sz-exp-0064-lbfgs-log_processed_bs_16.txt};
        \addplot+ [mark=none,line width=1pt]
        table[x expr={(\thisrowno{1})},y expr={log10(\thisrowno{3})},col sep=space]{batch-sz-exp-0064-lbfgs-log_processed_bs_32.txt};
        \addplot+ [mark=none,line width=1pt]
        table[x expr={(\thisrowno{1})},y expr={log10(\thisrowno{3})},col sep=space]{batch-sz-exp-0064-lbfgs-log_processed_bs_64.txt};
        \legend{\scriptsize{BS=8}, \scriptsize{BS=16}, \scriptsize{BS=32}, \scriptsize{BS=64}}
      \end{axis}
    \end{tikzpicture}
    \subcaption{L-BFGS}
    \label{fig:64x64-lbfgs-loss-epoch}
    \end{center}
    \end{minipage}
  \caption{\ipoint{Loss vs. epoch for $64\times 64$ domain DiffNet with varying batch-size (BS). Second order optimizers (L-BFGS) provide minimal convergence drift with higher batch-size. All training done on 1 node of Nova}}
    \label{fig:loss-epoch-plots-64x64}
\end{figure}

\ipoint{\textbf{Effect of batch size on solution using first order (SGD) and second order (L-BFGS) optimizers:} The data parallelism afforded by DeepFusion potentially allows one to use larger batch sizes. However, it is well known that increasing batch sizes can decrease the convergence of Stochastic Gradient Descent (SGD). We explore this effect of batch size on the training performance. Fig.~\ref{fig:loss-epoch-plots-64x64}(top) plots evolution of the loss function with training epochs for increasing batch size (BS) for a $64 \times 64$ resolution DiffNet. We clearly see some degradation in convergence rate as the batch size is increased to 64. We next trained the same network using a second order optimizer (L-BFGS) implemented in the DeepFusion framework. Fig~\ref{fig:loss-epoch-plots-64x64}(bottom) shows negligible impact of increasing batch size on convergence. As expected, the second order method converges in fewer epochs, with similar reduction in loss happening within 10 training epochs as compared to 150 epochs for SGD. L-BFGS optimizers require larger memory (to evaluate the Hessian) which a distributed approach (like DeepFusion or LBANN) can gracefully accommodate.}

\begin{figure}[!ht]
    \centering
    \begin{minipage}{.49\textwidth}
  \begin{center}
  \begin{tikzpicture}
      \begin{axis}[
          width=0.99\linewidth, 
          height=4cm,
          xlabel={time (sec)}, 
          ylabel=$\log_{10}(L)$, 
          legend style={at={(0.95,0.95)},anchor=north east,legend columns=2}, 
          ymin=-0.7,
          ymax=1.5,
          grid=both,
          x tick label style={rotate=0,anchor=north}
        ]
        \addplot+ [mark=none,line width=1pt]
        table[x expr={(\thisrowno{0})},y expr={log10(\thisrowno{3})},col sep=space]{batch-sz-exp-0064-sgd-log_processed_bs_8.txt};
        \addplot+ [mark=none,line width=1pt]
        table[x expr={(\thisrowno{0})},y expr={log10(\thisrowno{3})},col sep=space]{batch-sz-exp-0064-sgd-log_processed_bs_16.txt};
        \addplot+ [mark=none,line width=1pt]
        table[x expr={(\thisrowno{0})},y expr={log10(\thisrowno{3})},col sep=space]{batch-sz-exp-0064-sgd-log_processed_bs_32.txt};
        \addplot+ [mark=none,line width=1pt]
        table[x expr={(\thisrowno{0})},y expr={log10(\thisrowno{3})},col sep=space]{batch-sz-exp-0064-sgd-log_processed_bs_64.txt};
        \legend{\scriptsize{BS=8}, \scriptsize{BS=16}, \scriptsize{BS=32}, \scriptsize{BS=64}}
      \end{axis}
    \end{tikzpicture}
    \subcaption{SGD}
    \label{fig:64x64-sgd-loss-time}
    \end{center}
    \end{minipage}
    \\
    \begin{minipage}{.49\textwidth}
  \begin{center}
  \begin{tikzpicture}
      \begin{axis}[
          width=0.99\linewidth, 
          height=4cm,
          xlabel={time (sec)}, 
          ylabel=$\log_{10}(L)$, 
          legend style={at={(0.95,0.95)},anchor=north east,legend columns=2}, 
          ymin=-0.7,
          ymax=1.5,
          grid=both,
          x tick label style={rotate=0,anchor=north}
        ]
        \addplot+ [mark=none,line width=1pt]
        table[x expr={(\thisrowno{0})},y expr={log10(\thisrowno{3})},col sep=space]{batch-sz-exp-0064-lbfgs-log_processed_bs_8.txt};
        \addplot+ [mark=none,line width=1pt]
        table[x expr={(\thisrowno{0})},y expr={log10(\thisrowno{3})},col sep=space]{batch-sz-exp-0064-lbfgs-log_processed_bs_16.txt};
        \addplot+ [mark=none,line width=1pt]
        table[x expr={(\thisrowno{0})},y expr={log10(\thisrowno{3})},col sep=space]{batch-sz-exp-0064-lbfgs-log_processed_bs_32.txt};
        \addplot+ [mark=none,line width=1pt]
        table[x expr={(\thisrowno{0})},y expr={log10(\thisrowno{3})},col sep=space]{batch-sz-exp-0064-lbfgs-log_processed_bs_64.txt};
        \legend{\scriptsize{BS=8}, \scriptsize{BS=16}, \scriptsize{BS=32}, \scriptsize{BS=64}}
      \end{axis}
    \end{tikzpicture}
    \subcaption{L-BFGS}
    \label{fig:64x64-lbfgs-loss-time}
    \end{center}
    \end{minipage}
      \caption{\ipoint{Loss vs. time to solve for $64\times 64$ domain DiffNet with varying batch-size (BS). Second order optimizers (L-BFGS) are substantially slower than SGD. All training done on 1 node of Nova}}
    \label{fig:loss-time-plots-64x64}
\end{figure}
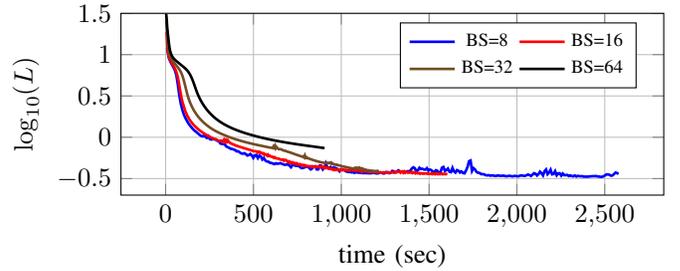
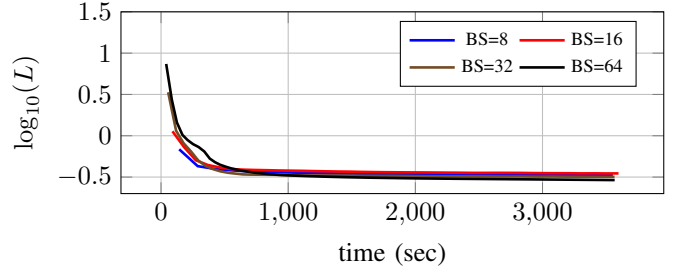

\ipoint{The increased computational overhead results in increased time per epoch, and this is plotted in Fig.~\ref{fig:loss-time-plots-64x64}. While L-BFGS takes an order of magnitude less number of epochs to converge, each epoch is more expensive due to memory and compute requirements from evaluating the Hessian. It is informative also to look at the results in terms of computational time to reach a certain convergence threshold, with L-BFGS schemes about $3 \times$ faster than the SGD scheme. This reduction in computational time to reach a desired loss threshold can be enhanced via parallelization. DeepFusion allows parallelization of L-BFGS based training across multiple CPU nodes, and this training across multiple nodes proportionally decreases the time-to-train, as shown in Table~\ref{tab:128x128-lbfgs-scaling-nova}}

\begin{table}[]
    \centering
    \ipoint{
\csvreader[
  tabular=|c|c|c|,
  table head=\hline 
  \bfseries{\boldmath${p}$} & \bfseries{\boldmath${64 \times 64}$} & \bfseries{\boldmath${128 \times 128}$} 
  \\ \hline,
  late after last line=\\ \hline 
]{
  lbfgs_100_epochs_combined_064_128_minute.txt
}{}{\csvlinetotablerow}
}
    \caption{\ipoint{Time (in minutes) taken to complete $100$ epochs of the L-BFGS method for the $64\times 64$ and $128\times 128$ domain DiffNet on different number of processors}}
    \label{tab:128x128-lbfgs-scaling-nova} \vspace{-0.2in}
\end{table}

\ipoint{These results are very promising as they allow using second order methods -- which are less sensitive to large batch sizes -- and data parallelization to reduce time-to-solve to beat SGD based approaches. We include additional results in the appendix A.}

{\bf Comparison of run-times between DiffNet and FEM based solutions}: While the training times for ML models is admittedly long, once trained, the inference step is often very fast. This allows one to amortize the cost of training across multiple users and instances. The availability of a general 'NeuralPDE' makes this a viable possibility. We quantify this argument by comparing the time it takes for a trained DiffNet to make a prediction (i.e. inference) with the time it takes for a well optimized FEM solver to perform the same prediction. \ipoint{Both DiffNet inference and FEM solve are performed on one node of Stampede2.} This comparison is reported in Table~\ref{tab:comparison_FEM_ML}, which shows a 40x improvement in prediction time. We emphasize that the inference step is not optimized, suggesting that the 40x improvement we show is a lower bound. 

\begin{table}[!ht]
  \begin{center}
    \begin{tabular}{|c|c|c|c|}
    \hline
     {\bf Domain Size} &   {\bf FEM (seconds)} & {\bf DeepFusion (seconds)} \\
     \hline
     512$\times$512	& 23.2 & 3.6 \\
     \hline
     1024$\times$1024 & 395.6 & 9.8 \\
     \hline
\end{tabular}
\caption{Comparison of solve time for the finite element solution  and the inference time for the DeepFusion solution \label{tab:comparison_FEM_ML}}
\end{center}
\end{table}

\vspace{-0.2in}
\section{Conclusions}
In this work, we report on a data distributed computing approach for training large neural network architectures, especially in the context of data-free generative models that serve as PDE solvers. We  highlighted  some of the key challenges and improvements over conventional GPU based training strategies, as well as other data-parallel approaches. We demonstrated excellent scaling  results  for  our  framework  on current  supercomputers. We illustrated the ability of this framework to enable practitioners to train very large models, thus enabling practical applications of such 'neuralPDE' solvers. We  believe  that availability of tools like the one presented here will help democratize the ability of a data scientist to produce (near) real time predictions of complex systems characterized by PDEs. Our future goals include extension of the framework to incorporate model parallelism for increased scaleup, as well as apply second order strategies to train DiffNets for a wide range of PDE's. 

\clearpage


\clearpage

\section{Acknowledgements}
We acknowledge support from ARPA-E DIFFERENTIATE program (DE-AR0001215), and computing support from XSEDE and Iowa State University. BG and BK also acknowledge partial support from NSF 1935255.

\bibliographystyle{IEEEtran}
\bibliography{ms}

\clearpage
\section{Appendix}

\subsection{Batch size effect on convergence}
\begin{figure}[!ht]
    \centering
    \begin{minipage}{.4\textwidth}
  \begin{center}
  \begin{tikzpicture}
      \begin{axis}[
          width=0.99\linewidth, 
          height=4cm,
          xlabel={Epoch}, 
          ylabel=$\log_{10}(L)$, 
          legend style={at={(0.95,0.95)},anchor=north east,legend columns=2}, 
          xmin=0,
          xmax=310,
          ymin=-1.5,
          ymax=1.0,
          grid=both,
          x tick label style={rotate=0,anchor=north}
        ]
        \addplot+ [mark=none,line width=1pt]
        table[x expr={(\thisrowno{1})},y expr={log10(\thisrowno{3})},col sep=space]{batch-sz-exp-0128-sgd-log_processed_bs_8.txt};
        \addplot+ [mark=none,line width=1pt]
        table[x expr={(\thisrowno{1})},y expr={log10(\thisrowno{3})},col sep=space]{batch-sz-exp-0128-sgd-log_processed_bs_16.txt};
        \addplot+ [mark=none,line width=1pt]
        table[x expr={(\thisrowno{1})},y expr={log10(\thisrowno{3})},col sep=space]{batch-sz-exp-0128-sgd-log_processed_bs_32.txt};
        \addplot+ [mark=none,line width=1pt]
        table[x expr={(\thisrowno{1})},y expr={log10(\thisrowno{3})},col sep=space]{batch-sz-exp-0128-sgd-log_processed_bs_64.txt};
        \legend{\scriptsize{BS=8}, \scriptsize{BS=16}, \scriptsize{BS=32}, \scriptsize{BS=64}}
      \end{axis}
    \end{tikzpicture}
    \subcaption{SGD}
    \label{fig:128x128-sgd-loss-epoch}
    \end{center}
    \end{minipage}
    \vskip 8pt
\begin{minipage}{.4\textwidth}
  \begin{center}
  \begin{tikzpicture}
      \begin{axis}[
          width=0.99\linewidth, 
          height=4cm,
          xlabel={Epoch}, 
          ylabel=$\log_{10}(L)$, 
          legend style={at={(0.95,0.95)},anchor=north east,legend columns=2}, 
          xmin=0,
          xmax=200,
          ymin=-1.5,
          ymax=1.0,
          grid=both,
          x tick label style={rotate=0,anchor=north}
        ]
        \addplot+ [mark=none,line width=1pt]
        table[x expr={(\thisrowno{1})},y expr={log10(\thisrowno{3})},col sep=space]{batch-sz-exp-0128-lbfgs-log_processed_bs_8.txt};
        \addplot+ [mark=none,line width=1pt]
        table[x expr={(\thisrowno{1})},y expr={log10(\thisrowno{3})},col sep=space]{batch-sz-exp-0128-lbfgs-log_processed_bs_16.txt};
        \addplot+ [mark=none,line width=1pt]
        table[x expr={(\thisrowno{1})},y expr={log10(\thisrowno{3})},col sep=space]{batch-sz-exp-0128-lbfgs-log_processed_bs_32.txt};
        \addplot+ [mark=none,line width=1pt]
        table[x expr={(\thisrowno{1})},y expr={log10(\thisrowno{3})},col sep=space]{batch-sz-exp-0128-lbfgs-log_processed_bs_64.txt};
        \legend{\scriptsize{BS=8}, \scriptsize{BS=16}, \scriptsize{BS=32}, \scriptsize{BS=64}}
      \end{axis}
    \end{tikzpicture}
    \subcaption{L-BFGS}
    \label{fig:128x128-lbfgs-loss-epoch}
    \end{center}
    \end{minipage}
    \caption{\ipoint{Loss vs. epoch for $128\times 128$ domain DiffNet with varying batch-size (BS). Second order optimizers (L-BFGS) show faster convergence rate and provide minimal convergence drift with higher batch-size.}}
    \label{fig:loss-epoch-plots-128x128}
\end{figure}
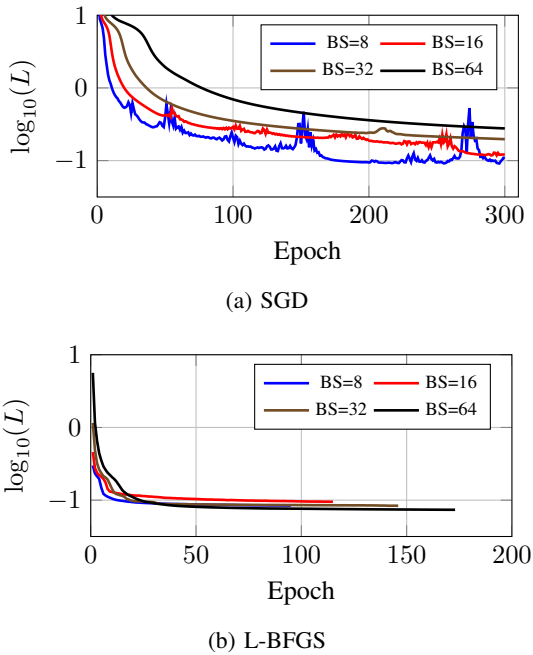

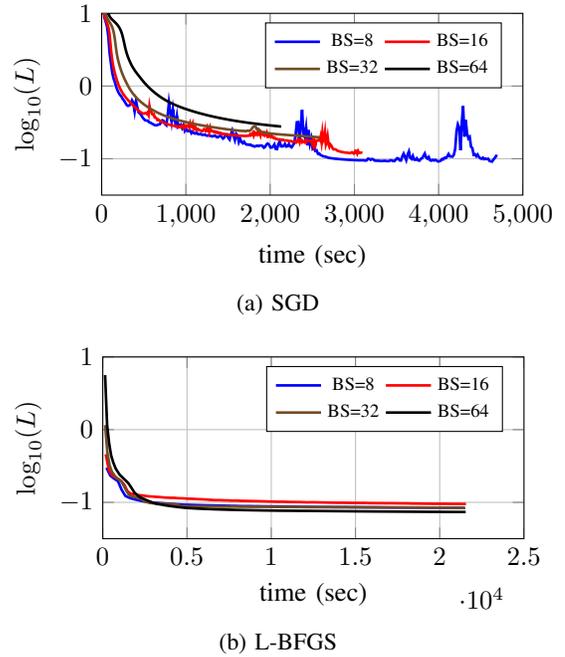
\begin{figure}[!ht]
    \centering
    \begin{minipage}{.4\textwidth}
  \begin{center}
  \begin{tikzpicture}
      \begin{axis}[
          width=0.99\linewidth, 
          height=4cm,
          xlabel={time (sec)}, 
          ylabel=$\log_{10}(L)$, 
          legend style={at={(0.95,0.95)},anchor=north east,legend columns=2}, 
          xmin=0,
          xmax=5000,
          ymin=-1.5,
          ymax=1.0,
          grid=both,
          x tick label style={rotate=0,anchor=north}
        ]
        \addplot+ [mark=none,line width=1pt]
        table[x expr={(\thisrowno{0})},y expr={log10(\thisrowno{3})},col sep=space]{batch-sz-exp-0128-sgd-log_processed_bs_8.txt};
        \addplot+ [mark=none,line width=1pt]
        table[x expr={(\thisrowno{0})},y expr={log10(\thisrowno{3})},col sep=space]{batch-sz-exp-0128-sgd-log_processed_bs_16.txt};
        \addplot+ [mark=none,line width=1pt]
        table[x expr={(\thisrowno{0})},y expr={log10(\thisrowno{3})},col sep=space]{batch-sz-exp-0128-sgd-log_processed_bs_32.txt};
        \addplot+ [mark=none,line width=1pt]
        table[x expr={(\thisrowno{0})},y expr={log10(\thisrowno{3})},col sep=space]{batch-sz-exp-0128-sgd-log_processed_bs_64.txt};
        \legend{\scriptsize{BS=8}, \scriptsize{BS=16}, \scriptsize{BS=32}, \scriptsize{BS=64}}
      \end{axis}
    \end{tikzpicture}
    \subcaption{SGD}
    \label{fig:128x128-sgd-loss-time}
    \end{center}
    \end{minipage}
    \vskip 8pt
    \begin{minipage}{.4\textwidth}
  \begin{center}
  \begin{tikzpicture}
      \begin{axis}[
          width=0.99\linewidth, 
          height=4cm,
          xlabel={time (sec)}, 
          ylabel=$\log_{10}(L)$, 
          legend style={at={(0.95,0.95)},anchor=north east,legend columns=2}, 
          xmin=0,
          xmax=25000,
          ymin=-1.5,
          ymax=1.0,
          grid=both,
          x tick label style={rotate=0,anchor=north}
        ]
        \addplot+ [mark=none,line width=1pt]
        table[x expr={(\thisrowno{0})},y expr={log10(\thisrowno{3})},col sep=space]{batch-sz-exp-0128-lbfgs-log_processed_bs_8.txt};
        \addplot+ [mark=none,line width=1pt]
        table[x expr={(\thisrowno{0})},y expr={log10(\thisrowno{3})},col sep=space]{batch-sz-exp-0128-lbfgs-log_processed_bs_16.txt};
        \addplot+ [mark=none,line width=1pt]
        table[x expr={(\thisrowno{0})},y expr={log10(\thisrowno{3})},col sep=space]{batch-sz-exp-0128-lbfgs-log_processed_bs_32.txt};
        \addplot+ [mark=none,line width=1pt]
        table[x expr={(\thisrowno{0})},y expr={log10(\thisrowno{3})},col sep=space]{batch-sz-exp-0128-lbfgs-log_processed_bs_64.txt};
        \legend{\scriptsize{BS=8}, \scriptsize{BS=16}, \scriptsize{BS=32}, \scriptsize{BS=64}}
      \end{axis}
    \end{tikzpicture}
    \subcaption{L-BFGS}
    \label{fig:128x128-lbfgs-loss-time}
    \end{center}
    \end{minipage}
    \caption{\ipoint{Loss vs. time for $128\times 128$ domain DiffNet shown in Fig.~\ref{fig:loss-epoch-plots-128x128}}}
    \label{fig:loss-time-plots-128x128}
\end{figure}

\textcolor{black}{We continue exploring the effect of batch-size on convergence rates of both first order optimizers (SGD) as well as second order optimizers (L-BFGS) both of which are implemented in a data parallel way in DeepFusion. Fig.~\ref{fig:loss-epoch-plots-128x128} plots the loss evolution with training epochs for a $128 \times 128$ DiffNet model that was trained on 1 node on Nova. These results are consistent with those shown in Fig.~\ref{fig:loss-epoch-plots-64x64} for a $64 \times 64$ resolution DiffNet model, where L-BFGS optimizer is relatively insensitive to the batch-size ranges chosen. As expected, the second order method converges in fewer epochs, with similar reduction in loss happening within 30 training epochs as compared to 150 epochs for SGD. L-BFGS optimizers require larger memory (to evaluate the Hessian) which a distributed approach (like DeepFusion) can gracefully accommodate.}
\textcolor{black}{The increased computational overhead results in increased time per epoch, and this is plotted in  Fig.~\ref{fig:loss-time-plots-128x128}.}

\vfill\break

\clearpage
\subsection{Additional quantitative comparison between DiffNet and conventional PDE solvers}
\begin{figure}[!ht]
    \centering
    \begin{minipage}[t]{.33\textwidth}
      \centering
      \begin{tikzpicture}
      \begin{axis}[
          width=0.99\linewidth, 
          xlabel=$x$, 
          ylabel=$u$,
          legend cell align={left},
          legend style={at={(0.98,0.02)},anchor=south east,legend columns=3}, 
          xmin=-0.1,
          xmax=1.1,
        grid=both,
          x tick label style={rotate=0,anchor=north} 
        ]
        \addplot+ [color=black,mark=none,line width=1.2pt]
        table[x expr={\thisrow{"arc_length"}},y expr={\thisrow{"u_ml"}},col sep=comma]{n_512_c_3.csv};
        \addplot+ [color=red,mark=none]
        table[x expr={\thisrow{"arc_length"}},y expr={\thisrow{"u_fem"}},col sep=comma]{n_512_c_3.csv};
        \addplot+ [color=blue,dashed,mark=none,line width=1pt]
        table[x expr={\thisrow{"arc_length"}},y expr={\thisrow{"u_fdm"}},col sep=comma]{n_512_c_3.csv};
        \legend{\tiny{SciDDL}, \tiny{FEM},\tiny{FDM}}
      \end{axis}
    \end{tikzpicture}
    \end{minipage}
    \\
    \begin{minipage}[t]{.33\textwidth}
      \centering
      \begin{tikzpicture}
      \begin{axis}[
          width=0.99\linewidth, 
          xlabel=$x$, 
          ylabel=$u$,
          legend cell align={left},
          legend style={at={(0.98,0.02)},anchor=south east,legend columns=3}, 
          xmin=-0.1,
          xmax=1.1,
        grid=both,
          x tick label style={rotate=0,anchor=north} 
        ]
        \addplot+ [color=black,mark=none,line width=1.2pt]
        table[x expr={\thisrow{"arc_length"}},y expr={\thisrow{"u_ml"}},col sep=comma]{n_512_c_5.csv};
        \addplot+ [color=red,mark=none]
        table[x expr={\thisrow{"arc_length"}},y expr={\thisrow{"u_fem"}},col sep=comma]{n_512_c_5.csv};
        \addplot+ [color=blue,dashed,mark=none,line width=1pt]
        table[x expr={\thisrow{"arc_length"}},y expr={\thisrow{"u_fdm"}},col sep=comma]{n_512_c_5.csv};
        \legend{\tiny{SciDDL}, \tiny{FEM},\tiny{FDM}}
      \end{axis}
    \end{tikzpicture}
    \end{minipage}
    \begin{minipage}[t]{.33\textwidth}
      \centering
      \begin{tikzpicture}
      \begin{axis}[
          width=0.99\linewidth, 
          xlabel=$x$, 
          ylabel=$u$,
          legend cell align={left},
          legend style={at={(0.98,0.02)},anchor=south east,legend columns=1}, 
          xmin=-0.1,
          xmax=1.1,
        grid=both,
          x tick label style={rotate=0,anchor=north} 
        ]
        \addplot+ [color=black,mark=none,line width=1.2pt]
        table[x expr={\thisrow{"arc_length"}},y expr={\thisrow{"u_ml"}},col sep=comma]{n_512_c_13.csv};
        \addplot+ [color=red,mark=none]
        table[x expr={\thisrow{"arc_length"}},y expr={\thisrow{"u_fem"}},col sep=comma]{n_512_c_13.csv};
        \addplot+ [color=blue,dashed,mark=none,line width=1pt]
        table[x expr={\thisrow{"arc_length"}},y expr={\thisrow{"u_fdm"}},col sep=comma]{n_512_c_13.csv};
        \legend{\tiny{SciDDL}, \tiny{FEM},\tiny{FDM}}
      \end{axis}
    \end{tikzpicture}
    \end{minipage}
    \hspace{0.1mm}
    \caption{\ipoint{The solution profile $u$ vs. $x$ at $t = 0.2$, compared for different methods. Resolution = $512\times 512$ and $c$ values are $3$ (top), $5$ (middle) and $13$ (bottom)}}
    \label{fig:plot-solutions-vs-x-t-1-resolution-512}
\end{figure}
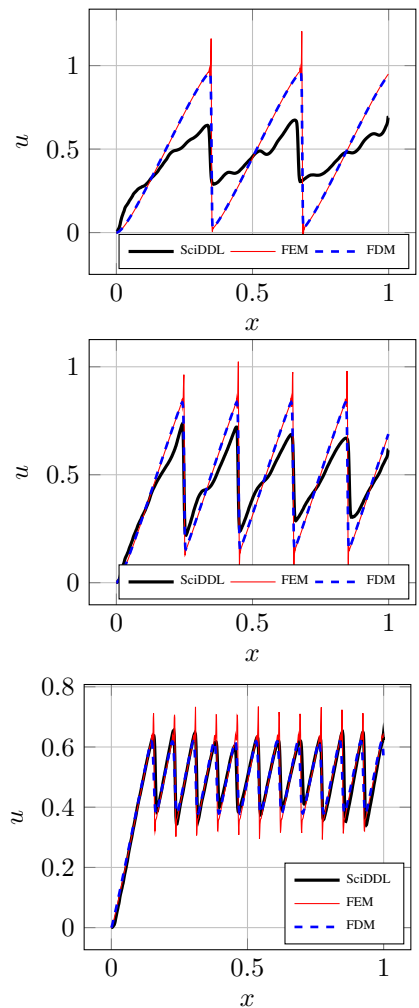

\textcolor{black}{In this subsection, we provide additional results over those shown in the main text to quantitatively compare the DiffNet inferences with conventional PDE solver technology. Figure~\ref{fig:plot-solutions-vs-x-t-1-resolution-512} plots the solution at a particular time instance ($t = 0.2$) where there is formation of shocks. Notice that the DiffNet solution is very close to the fully resolved FDM solution for the large wave-number case ($c= 13$), with increasingly large deviations as the wave-number is decreased.}

\end{document}